\documentclass{article}

\usepackage{microtype}
\usepackage{graphicx}
\usepackage{subfigure}
\usepackage{booktabs}

\usepackage{hyperref}
\usepackage[table]{xcolor}
\usepackage{makecell}
\usepackage{enumitem}
\usepackage{xspace}
\usepackage{ragged2e}
\usepackage{wrapfig}
\usepackage{pifont}
\newcommand{\cmark}{\ding{51}}
\newcommand{\xmark}{\ding{55}}
\usepackage{algorithm}
\usepackage{algorithmic}

\newcommand{\methodname}{PQS\xspace}
\newcommand{\modulenameone}{QSS\xspace}
\newcommand{\modulenametwo}{CSS\xspace}

\usepackage{amsmath}
\usepackage{amssymb}
\usepackage{mathtools}
\usepackage{amsthm}

\def\refine{\textcolor{black}}

\def\refinedv1{\textcolor{black}}

\def\0{{\bf 0}}
\def\1{{\bf 1}}

\newcommand{\ie}{\textit{i}.\textit{e}.}
\newcommand{\eg}{\textit{e}.\textit{g}.}

\newcommand{\RETURN}{\textbf{Return:}}

\usepackage{threeparttable}
\usepackage{url}
\usepackage{svg}
\svgpath{{./figures/}}

\usepackage[accepted]{icml2025}

\usepackage{amsmath}
\usepackage{amssymb}
\usepackage{mathtools}
\usepackage{amsthm}

\usepackage[capitalize,noabbrev]{cleveref}

\theoremstyle{plain}

\theoremstyle{definition}

\theoremstyle{remark}

\icmltitlerunning{Deep Electromagnetic Structure Design Under Limited Evaluation Budgets}

\begin{document}

\twocolumn[
\icmltitle{Deep Electromagnetic Structure Design Under Limited Evaluation Budgets}

\begin{icmlauthorlist}
\icmlauthor{Shijian Zheng}{scutft,pcl}
\icmlauthor{Fangxiao Jin}{scutft}
\icmlauthor{Shuhai Zhang}{scutse,pzl}
\icmlauthor{Quan Xue}{scutm}
\icmlauthor{Mingkui Tan}{scutse}
\end{icmlauthorlist}

\icmlaffiliation{scutft}{School of Future Technologies, South China University of Technology}
\icmlaffiliation{scutse}{School of Software Engineering, South China University of Technology}
\icmlaffiliation{scutm}{School of Microelectronics, South China University of Technology}

\icmlaffiliation{pzl}{Pazhou Laboratory}
\icmlaffiliation{pcl}{Peng Cheng Laboratory}
\icmlcorrespondingauthor{Mingkui Tan}{mingkuitan@scut.edu.cn}
\icmlcorrespondingauthor{Quan Xue}{eeqxue@scut.edu.cn}

\icmlkeywords{Machine Learning, ICML}

\vskip 0.3in
]

\printAffiliationsAndNotice{}

\begin{abstract}

Electromagnetic structure (EMS) design plays a critical role in developing advanced antennas and materials, but remains challenging due to high-dimensional design spaces and expensive evaluations.
While existing methods commonly employ high-quality predictors or generators to alleviate evaluations, they are often data-intensive and struggle with real-world scale and budget constraints.
To address this, we propose a novel method called Progressive Quadtree-based Search (\methodname).
Rather than exhaustively exploring the high-dimensional space, \methodname converts the conventional image-like layout into a quadtree-based hierarchical representation, enabling a progressive search from global patterns to local details.
Furthermore, to lessen reliance on highly accurate predictors, we introduce a consistency-driven sample selection mechanism. This mechanism quantifies the reliability of predictions, balancing exploitation and exploration when selecting candidate designs.
We evaluate \methodname on two real-world engineering tasks, \ie, Dual-layer Frequency Selective Surface and High-gain Antenna. Experimental results show that our method can achieve satisfactory designs under limited computational budgets, outperforming baseline methods. In particular, compared to generative approaches, it cuts evaluation costs by 75$\sim$85\%, effectively saving \refine{20.27$\sim$38.80} days of product designing cycle.

\end{abstract}

\section{Introduction}

\begin{figure}[t]

  \centering
   \includegraphics[width=1\linewidth]{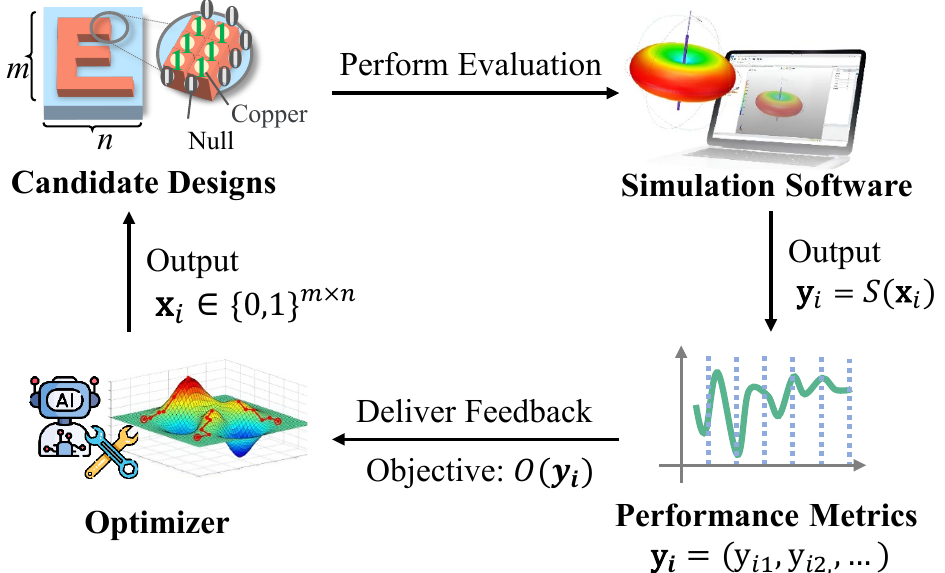}
    \vspace{-0.3in}

    \caption{Illustration of the EMS design workflow.
    }
  \label{fig:problem_defin}
  \vspace{-0.25in}
\end{figure}
Electromagnetic structure (EMS) is designed to interact with electromagnetic waves, which is crucial for various domains, ranging from telecommunications to 5G antennas, including frequency-selective surface~\citep{zhu2022adversarial}, metamaterials ~\citep{chen2023correlating, deng2021benchmarking}, photonic crystals~\citep{peurifoy2018nanophotonic}, and circuit \citep{cheng2022policy, shahane2023graph}.
Despite its broad range of applications, EMS design is challenging primarily for two major reasons.

Firstly, EMS design involves a \emph{vast design space}. For instance, modeling the structure as a $12\times24$ grid with two possible states (metal or empty) per cell, one obtains $b^d = 2^{288} \approx 10^{86}$ possible configurations. The sheer magnitude of this combinatorial space poses a formidable challenge for both domain experts and algorithms, making it difficult to locate promising designs in an efficient manner.

Secondly, evaluations are \emph{costly}. Evaluating an EMS design often relies on time-consuming, full-wave electromagnetic simulations, typically governed by complex partial differential equations~\citep{koziel2014antenna}, which cannot be replaced by simpler analytical approximations. According to a technical report from Inceptra\footnote{\href{https://www.inceptra.com/how-computer-hardware-impacts-cst-electromagnetic-simulation-speed/}{https://www.inceptra.com/how-computer-hardware-impacts-cst-electromagnetic-simulation-speed/}}, simulating a single design may take anywhere between 660 and 42,780 seconds.

Recent works have extended techniques from other high-dimensional, expensive design domains to EMS,
typically categorized into two major directions.
\textbf{Predictor-based methods}~\citep{koziel2022low-cost, jing2022neural}
seek to alleviate expensive evaluations by training a deep neural network (DNN) surrogate to approximate costly electromagnetic simulations.
Once trained, this surrogate can be integrated into standard optimizers (e.g., gradient-based or evolutionary algorithms),
thus reducing reliance on full-wave solvers.
Meanwhile, \textbf{generative approaches}~\citep{brookes2019conditioning, Gao2023Pifold}
focus on directly generating promising design candidates under certain performance constraints,
often through conditional generative models.
By learning an end-to-end mapping from performance targets to structural layouts,
these methods aim to bypass exhaustive searches over the vast design space.
\emph{Unfortunately}, both predictor-based and generative approaches still exhibit significant drawbacks when scaled to real-world EMS tasks.

On the one hand, existing methods typically operate directly in a high-dimensional design space,
treating \emph{every pixel} as a free variable to be optimized.
As the layout size increases, the optimization process can incur rapidly growing evaluation costs,
 which is commonly known as the curse of dimensionality \citep{chen2015measuring}.
In practical EMS scenarios, where each simulation is time-consuming, such high-dimensional exploration quickly becomes infeasible.

On the other hand, many recent advances heavily rely on \emph{high-fidelity models},
which merely shift the cost from direct full-wave evaluations to extensive data collection.
Training such models for EMS tasks often demands tens of thousands of labeled samples,
as exemplified by the 10,000--20,000 samples in \citet{wang2023end} and the 20,000--2,000,000 in \citet{majorel2022deep}.
Acquiring these large datasets can take months or even years of simulation time, making such solutions impractical
for industry applications that must adhere to strict deadlines and budget constraints.

\begin{table}[t]
  \vspace{-0.1in}
\label{tab:mbo}
  \centering
  \caption{Comparison of different structure design tasks.}

  \rowcolors{2}{gray!15}{white}

  \resizebox{1\linewidth}{!}{

\begin{tabular}{l!{\color{black}\vrule}lll}
    \toprule
      & NAS
      & Molecule Design
      & EMS Design \\
    \midrule
     Large Design Space
      & \cmark\ $10^{4}\sim10^{18}$
      & \cmark\ $10^{6}$
      & \cmark\ $10^{86}\sim10^{90}$ \\
     High Evaluation Cost

      & \multicolumn{3}{c}{\cmark\ Ranging from minutes to days} \\
    Public Dataset
      & \cmark\ NAS-bench
      & \cmark\ ChEMBL
      & \xmark \\
     \makecell[l]{Pretrained \\ Predictor or Generator}
      & \cmark\ NAS-bench-301
      & \cmark\ AlphaFold
      & \xmark \\
    Data Augmentation
      & \cmark\ Weight Sharing
      & \cmark\ AugLiChem
      & \xmark \\
    \bottomrule
    \end{tabular}
  }
    \vspace{-0.2in}
\end{table}

To overcome the aforementioned challenges, we propose a novel Progressive Quadtree-based Search (\methodname) method.
Instead of treating every pixel in the layout as a free variable, \methodname introduces a \textbf{Quadtree-based Search Strategy} (\modulenameone). By employing a hierarchical quadtree representation, \modulenameone reduces the intractable complexity of exploring every pixel simultaneously and focuses computational resources on the most promising regions.
Moreover, to lessen reliance on high-quality predictors, \methodname introduces a \textbf{Consistency-based Sample Selection} (\modulenametwo) mechanism. \modulenametwo leverages a consistency metric to gauge the reliability of model outputs.
Based on this metric, \modulenametwo adapts its use of the limited evaluation budget. If predictions are consistent, more resources are devoted to promising designs. Conversely, the method assigns additional evaluations to designs currently deemed suboptimal, ensuring that potentially superior solutions are not overlooked.
Together, these two modules accelerate the discovery of high-quality designs, even under real-world large-scale tasks with tight simulation budgets.

Our contributions are summarized as follows:
\begin{itemize}[leftmargin=*]
\item \textbf{A novel Progressive Quadtree-based Search (\methodname) approach for EMS design.}
We present \methodname to tackle the dual challenge of high-dimensional design spaces and limited evaluation budgets.
By progressively narrowing down the design space and strategically allocating computational resources,
\methodname substantially reduces the reliance on massive datasets and repeated full-wave simulations,
making it well-suited for industrial scenarios with strict budget constraints.
\item \textbf{A Quadtree-based Search Strategy (\modulenameone) for hierarchical design exploration.}
Instead of treating each pixel as an independent variable, we introduce a hierarchical quadtree representation
that partitions the layout into coarser subregions first, and then refines promising areas in finer detail.
This strategy effectively mitigates the curse of dimensionality by focusing on higher-potential regions,
thereby accelerating optimization and reducing the overall simulation overhead.
\item \textbf{A Consistency-based Sample Selection (\modulenametwo) mechanism for adaptive resource allocation.}
To lessen the reliance on high-fidelity models, \modulenametwo measures the reliability of surrogate predictions
via a consistency metric. When predictions are deemed consistent, more evaluations target promising designs;
when predictions exhibit high uncertainty, additional true evaluations are triggered
to prevent overlooking potentially optimal solutions. This adaptive balance between exploration and exploitation
enhances search efficiency under limited evaluation budgets.
    \end{itemize}

\section{Related Work}

\subsection{Electromagnetic Structure Design}
Recent years have seen various data-driven approaches introduced to reduce the burden of expensive full-wave simulations.
For instance, surrogate models coupled with evolutionary or gradient-based optimization \citep{naseri2022synthesis,jia2023knowledge,zheng2023unifying} have demonstrated efficiency in moderate-scale EMS scenarios, but constructing accurate surrogates for large-scale spaces remains challenging.

Meanwhile, inverse design with generative models has also gained traction, such as cGAN \citep{an2021multifunctional} or cVAE \citep{lin2022machine}, enabling the direct generation of candidate layouts that meet certain performance criteria.

However, most existing approaches still require substantial numbers of samples to train high-fidelity models, thereby incurring considerable simulation expenses.

\subsection{Analogous Structure Design}
Beyond EMS, data-driven techniques have been extensively explored in domains such as neural architecture search (NAS) and molecule design, both of which also share the challenges of vast design spaces and expansive evaluation (see Table~\ref{tab:mbo}).
For NAS, DARTS \citep{liu2018darts} and PNAS \citep{liu2018progressive} iteratively refine candidate neural networks in a vast design space, with benchmarks such as NAS-Bench \citep{ying2019bench, dong2020bench} offering public datasets and pretrained predictors (e.g., NAS-Bench-301 \citep{siems2020bench}). Moreover, augmentation techniques such as weight sharing can derive approximate performance labels from the existing data\citep{ren2021comprehensive}. These available infrastructures reduce evaluation overhead.
Similarly, in molecule design, researchers leverage large-scale databases like ChEMBL \citep{gaulton2017chembl}, pre-trained models (\eg AlphaFold \citep{jumper2021highly}; GFlowNets \citep{madan2023learning}), and data augmentation strategies \citep{magar2022auglichem,han2019progan}.

In contrast, EMS design is burdened by a larger search space ($10^{86}\sim10^{90}$) and lacks public datasets, pre-trained models, or augmentation methods (see Table~\ref{tab:mbo}).
Moreover, EMS are typically developed in a highly customized manner in industrial scenarios, with diverse objectives and boundary conditions. This unique task setting forces researchers to \emph{design from scratch}, making EMS design fundamentally different from other structure design problems.

\section{Preliminary and Problem Statement}
To formulate EMS design problem, we first give some necessary definitions.

\textbf{Design Space}:
We consider a binary design space $\mathcal{X} \subseteq \{0, 1\}^{m \times n}$, where each design $\mathbf{x} \in \mathcal{X}$ is represented as an $m \times n$ matrix. The element $x_{ij}$ indicates a specific material (e.g., metal vs. empty) placed at grid $(i, j)$ in the layout.

\textbf{Performance Evaluation}:
To quantify the performance of a design $\mathbf{x}$, we employ a high-fidelity simulator $S$, which numerically solves the underlying Maxwell's equations~\citep{bondeson2012computational}. The simulator outputs a $p$-dimensional vector $\mathbf{y} = S(\mathbf{x}) = [y_{1}, \ldots, y_{p}]^\top$, where each component $y_{k} = S_k(\mathbf{x})$ is a performance criterion of interest.

\textbf{Optimization Formulation}:
In practical tasks, the design must deliver robust performance across multiple criteria simultaneously, rather than excelling in only one while sacrificing others. We thus define the objective function as the minimum performance among all criteria:
\begin{equation}
    O(S(\mathbf{x}))
    = \min_{1 \leq k \leq p} \, S_k(\mathbf{x}).
    \label{eq:robust_objective}
\end{equation}
In other words, $O(S(\mathbf{x}))$ captures the worst-case among the $p$ performance metrics. Maximizing this objective ensures that no single metric falls below an acceptable threshold.

We seek to find a design $\mathbf{x}$ that maximizes $O(S(\mathbf{x}))$ under a strict simulation budget $T_{\text{max}}$, yielding the following formulation:
\begin{equation}
    \max_{\mathbf{x} \in \mathcal{X}} \, O(S(\mathbf{x}))
    \quad
    \text{subject to}
    \quad
    T \le T_{\text{max}},
    \label{eq:constrained_surrogate_optimization}
\end{equation}
where $T$ is the number of full-wave simulations invoked during the optimization.

Since calling the simulator $S$ is computationally expensive, it is common to introduce an predictor $f_{\theta}$ to approximate $S(\mathbf{x})$. The details will be presented in Section~\ref{sec:methods}.

\section{Proposed Methods}
\label{sec:methods}
\begin{figure*}[ht]

\begin{minipage}[t]{0.47\linewidth}
    \centering
    \footnotesize
    \begin{algorithm}[H]\small
    \caption{General scheme of \methodname for EMS.}
    \label{algorithm}
 \begin{algorithmic}
    \REQUIRE{Initial dataset $D_0$, maximum of simulation runs $T_{\text{max}}$, maximum of tree nodes $N_\text{max}$}.
    \STATE Dataset $D \gets D_0$.
    \STATE Current runs of simulation $T \gets \text{length}(D_0)$.
    \WHILE{$T \leq T_{\text{max}}$}
    \STATE Train initial predictor $f_{\theta}$.
    \STATE Conduct optimization under Quadtree-based Search Strategy in algorithm~\ref{algorithm2}.
    \STATE Conduct Consistency-based Sample Selection based on Eqn.~(\ref{eqn:select}).
    \STATE Conduct Simulation and obtain feedback $\{(x, y)\}$.
    \STATE Add $\{(x, y)\}$ to dataset $D$.
    \STATE Update predictor $f_{\theta}$ using $D$.
            \STATE Update Current runs of simulation $T$.
    \ENDWHILE

\RETURN The satisfactory solutions from $\mathcal{D}$.
\end{algorithmic}
    \end{algorithm}
\end{minipage}\hspace{0.27cm}
\begin{minipage}[t]{0.49\linewidth}
    \centering
    \footnotesize
    \begin{algorithm}[H]\small
    \caption{EMS Optimization with \modulenameone.}
    \label{algorithm2}
\begin{algorithmic}
\REQUIRE Maximum of leaf nodes \( N_{\text{max}} \), predictor \( f_\theta \), size \( K \) of Top-K, Maximum iterations $M$.

\STATE Initialize the root node \( n_{\text{root}} \), leaf node set \(L\) and Top-K list.

\FOR{each $i \in [1,M]$}
\WHILE{$|L| < N_{\text{max}}$}
    \STATE Randomly select a leaf node \( n \) from \( L \).

    \STATE Resample node state \( s_n \sim \text{Uniform}(\{0,1\}) \) or split the node based on Eqn.~(\ref{eq:midpoint}).

    \STATE Reconstruct the design matrix \( \mathbf{x} \) based on Eqn.~(\ref{eq:matrix_reconstruction}) and evaluate \( O(f_\theta(\mathbf{x})) \).

        \STATE Update the Top-K list.

\ENDWHILE
\ENDFOR
\STATE Conduct Depth-wise Importance Assignment based on Eqn.~(\ref{eq:ia}).
\RETURN The Top-K best designs \( \{ \mathbf{x}_k^* \}_{k=1}^{K} \).
\end{algorithmic}
    \end{algorithm}
\end{minipage}
\vspace{-0.85em}
\end{figure*}

\begin{figure*}[t]

\centering
\includegraphics[width=1\linewidth]{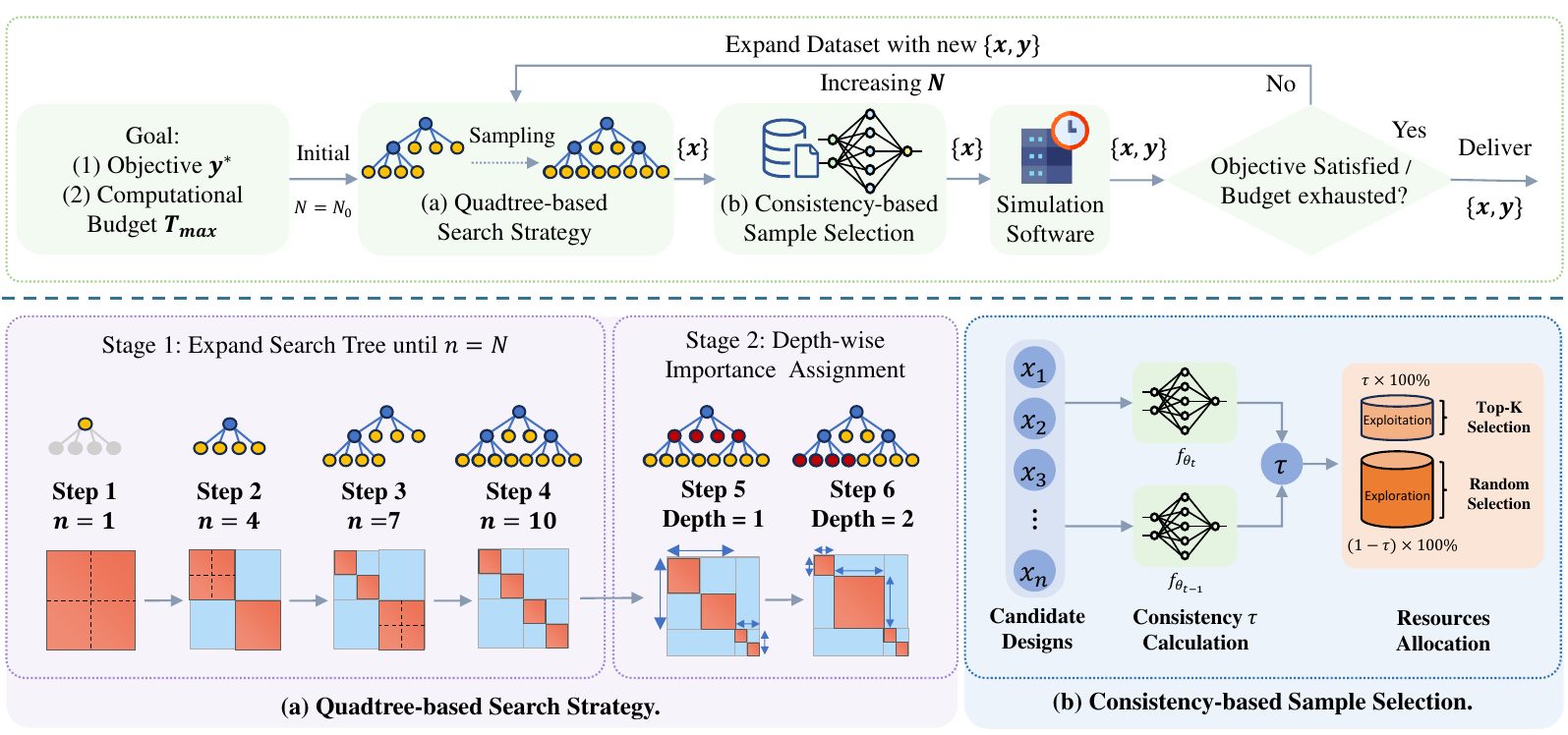}
\vspace{-0.2in}
\caption{\textbf{An illustration of the proposed \methodname}. The top portion illustrates the main workflow: (1)
the \emph{ s} generates candidate designs in a progressively refined design space;
(2) a predictor $f_{\theta_t}$ then approximates the performance of candidates;
(3) the \emph{Consistency-based Sample Selection} mechanism determines which designs proceed to high-fidelity simulations; (4) in the next iteration, the allowable number of leaf nodes $N$ grows to enable more complex designs, while $f_{\theta_t}$ is updated with newly evaluated samples.
The lower portion provide details of modules: (a)~\modulenameone proceeds in two stages, first performing a tree search to capture global patterns and then refining local details; (b) \modulenametwo uses the prediction consistency $\tau$ to balance between an exploitation or exploration strategy.
}
\vspace{-0.01in}
\label{fig:overall_illustration}
\end{figure*}

We propose \emph{Progressive Quadtree-based Search} (\methodname), to tackle the EMS design under limited evaluation budgets.
To this end, \methodname integrates two complementary modules: 1) \emph{Quadtree-based Search Strategy}, which employs a quadtree representation and a progressive refinement process to reduce the dimensionality of the design space, and 2) \emph{Consistency-
based Sample Selection}, which efficiently allocates simulation resources by leveraging imperfect predictor outputs, thus maximizing the benefits of limited evaluations. An overview of our framework is illustrated in Figure~\ref{fig:overall_illustration}. Algorithm~\ref{algorithm} summarizes the entire process. Below, we detail each component of \methodname.

\subsection{Quadtree-based Search Strategy}
\label{sec:Progressive}
Directly searching at the pixel level across an entire EMS layout can be computationally expensive. Therefore, we propose a Quadtree-based Search Strategy (\modulenameone), which leverages a  quadtree-based representation to reduce the scale of the design space and a progressive tree-search method to refine only those regions that demand higher resolution. The pseudo-code of \modulenameone is summarized in Algorithm~\ref{algorithm2}.

\paragraph{Quadtree-based Representation.}
In conventional pixel-based layouts, every cell is treated uniformly, which inflates the search space. Our idea starts from allowing for varying resolutions across different regions. To this end, we employ a quadtree to represent these regions. The quadtree structure recursively subdivides the layout into four child nodes when necessary. Homogeneous areas remain as leaf nodes, requiring minimal variables,
while more intricate areas get subdivided into smaller blocks, enabling finer resolution.

Specifically, each node \(n\) in the quadtree \(Q\) corresponds to a rectangular subregion of the layout matrix \(\mathbf{x}\), determined by the row indices
\(\bigl[r_n^{\text{start}},\,r_n^{\text{end}}\bigr]\)
and the column indices
\(\bigl[c_n^{\text{start}},\,c_n^{\text{end}}\bigr]\).
Formally, for any pixel \((i,j)\) within \(n\)'s subregion, we have
\begin{equation}
r_n^{\text{start}} \;\le\; i \;\le\; r_n^{\text{end}},
\quad
c_n^{\text{start}} \;\le\; j \;\le\; c_n^{\text{end}}.
\label{eq:range}
\end{equation}
If further refinement is needed, node \(n\) is split into four child nodes, each covering one of the four quadrants of the original subregion: upper-left \((n_0)\), upper-right \((n_1)\), lower-left \((n_2)\), and lower-right \((n_3)\). These child nodes recursively divide the region, with their row and column ranges determined by the midpoints of \(n\)'s intervals:
\begin{equation}
r_{\text{mid}} = \left\lfloor \frac{r_n^{\text{start}} + r_n^{\text{end}}}{2} \right\rfloor,
\quad
c_{\text{mid}} = \left\lfloor \frac{c_n^{\text{start}} + c_n^{\text{end}}}{2} \right\rfloor,
\label{eq:midpoint}
\end{equation}
so \refine{that}, for example, the upper-left child node \(n_0\) covers rows \([r_n^{\text{start}},\,r_{\text{mid}}]\) and columns \([c_n^{\text{start}},\,c_{\text{mid}}]\), with analogous adjustments for the other quadrants.

Otherwise, if no further splitting is performed, node \(n\) becomes a \emph{leaf node}. Each leaf node stores a single binary value \(s_n \in \{0,1\}\), indicating whether its entire subregion is labeled as 0 or 1. This representation  reduces redundancy by capturing homogeneous regions with just one bit.

As more nodes subdivide from the root, the quadtree grows to reflect progressively finer resolution. For each pixel \(x_{i,j}\) in the final layout, its value is determined by the unique leaf node \(n\) whose subregion contains \((i,j)\). The complete matrix \(\mathbf{x}\) can thus be reconstructed via:
\begin{equation}
x_{i,j} = \sum_{n \in L} s_n \cdot \mathbb{I}_n(i,j),
\label{eq:matrix_reconstruction}
\end{equation}
where \(L\) is the set of leaf nodes, and \(\mathbb{I}_n(i,j)\) is an indicator that is 1 if \((i,j)\) lies in node \(n\)'s subregion and 0 otherwise.

This recursive subdivision effectively compresses the matrix while enabling different resolutions in distinct regions.
Let \(L\) be the current set of leaf nodes.
Since each leaf node has two possible states (0 or 1), the original pixel-wise design space is simplified to :
\begin{equation}
\mathcal{S} = \left\{\, \mathbf{s} = \{ s_n \}_{n \in L} \;\bigm|\; s_n \in \{0,1\} \right\},
\end{equation}
whose size is \(2^{|L|}\).
Hence, by controlling how the quadtree expands (\ie, maximum of \(L\)), we can progressively manage the complexity of the design space in a structured manner.

\paragraph{Progressive Tree Search.}
 We propose a hierarchical search strategy, which starts from a simple design space and gradually increases the complexity of the design. By partitioning the design matrix into smaller subregions, the search process can adaptively explore finer subregions.

As shown in Figure~\ref{fig:overall_illustration}(a), the tree search process begins with the root node \( n_{\text{root}} \), which represents a sample in the most simplified design space. Subsequently, based on the current set of leaf nodes \( L \), the design matrix \( \mathbf{x} \) is reconstructed, and its performance \( O(f_\theta(\mathbf{x})) \) is evaluated by a predictor \( f_\theta \). The tree search process maintains the current Top-K best design matrices \( \mathbf{x}_k^* \) and their corresponding performance values \( O_k^* \), where \( O_k^* = O(f_\theta(\mathbf{x}_k^*)) \) represents the \( k \)-th best performance found so far.

At each iteration, a leaf node \( n \) is randomly selected from \( L \), and either its state \( s_n \) is resampled or is split into four child nodes with randomly initialized states, with both actions following a Bernoulli distribution with parameter 0.5. Resampling explores alternative configurations without expanding the design space, while splitting increases search granularity. The design matrix \( \mathbf{x} \) is reconstructed based on the results of these operations and evaluated by the predictor. If the new performance exceeds the current \( k \)-th best performance \( O_k^* \), the Top-K list is updated, replacing the \( k \)-th best design matrix \( \mathbf{x}_k^* \) with \( \mathbf{x} \), and recording the corresponding performance value \( O_k^* \leftarrow O(f_\theta(\mathbf{x})) \). The iteration repeats until the number of leaf nodes reaches the preset limit \( N_{\text{max}} \), at which point the algorithm terminates and returns the final Top-K best design matrices \( \mathbf{x}_k^* \).

\textbf{Depth-wise Importance Assignment}.
In the process of expanding the design space, the initial division of nodes assumes a uniform distribution, treating each newly created leaf node as having equal importance within the EMS matrix. However, certain regions may contribute more significantly to overall performance. To address this, we introduce a further refinement phase targeting the Top-\(k\) designs to better optimize the EMS structure.
This further optimization problem is formulated as:
\begin{equation}
\label{eq:ia}
 \max_{\mathbf{s'} \in \mathcal{S'}} O(f_\theta(\mathbf{x_{s'}})),
 \end{equation}
where \( \mathbf{x_{s'}} \) represents the EMS matrix defined by the quadtree structure, and \( \mathcal{S'} \) is the search space comprising all possible partition parameters in the quadtree:
\begin{equation}
\label{eq:ia2}
\mathcal{S'} = \{ (r_{n}^{\text{start}}, r_{n}^{\text{end}}, c_{n}^{\text{start}}, c_{n}^{\text{end}})\ \mid n\in Q \}.
\end{equation}
\subsection{Consistency-based Sample Selection}
\label{sec:ada}

After the \modulenameone module generates candidate samples, it is often infeasible to simulate all of them due to the high computational cost. Therefore, we must prioritize which samples to evaluate. To achieve this, we propose a \textbf{Consistency-based Sample Selection} (CSS) strategy, which optimizes the evaluation process to enhance search efficiency. The core steps of this strategy are as follows.

\textbf{Ranking-based Prediction Consistency.} Since our objective is to identify the best design, preserving \emph{ranking accuracy} is more critical than achieving precise numerical predictions. A model that correctly ranks candidates can still guide optimization effectively, even with moderate prediction errors. Therefore, we adopt \emph{Kendall’s tau}~\citep{kendall1938new} to measure the consistency between consecutive iterations’ predicted rankings. Specifically, at iteration~$t$, we compute $O(f_{\theta_t}(\mathbf{x}))$ for candidate designs $\{\mathbf{x}_i\}_{i=1}^n$ and compare these values to $O(f_{\theta_{t-1}}(\mathbf{x}))$ from iteration~$(t{-}1)$. The Kendall’s tau coefficient $\tau$ is given by:
	\begin{equation}\label{eqn:ktau} \begin{split}
\tau = \frac{2}{n(n-1)} \sum_{i<j} \text{sign}(O(f_{\theta_{t-1}}(\mathbf{x}_i)) - O(f_{\theta_{t-1}}(\mathbf{x}_j)))
\\
 \refinedv1{*}~\text{sign}(O(f_{\theta_t}(\mathbf{x}_i)) - O(f_{\theta_{t}}(\mathbf{x}_j))),
\end{split}
 \end{equation}
 where $ n $ is the number of data points, and $ x_i, x_j $ are the candidate samples. A value of $ \tau $ close to 1 indicates high consistency, while a value is close 0 or even negative indicates lower consistency.

\textbf{Mixed Selection.} When the model’s predictions become inaccurate, relying solely on the predictor’s recommended samples may overlook potentially better solutions.
To mitigate such bias, we introduce randomness for additional exploration.
Concretely, let $\tau$ be the Kendall’s tau coefficient from the preceding step.
A fraction $\tau$ of the total $R$ samples is chosen based on the predictor’s highest-ranked candidates,
while the remaining $1-\tau$ fraction is selected randomly.
Formally, we define:
\begin{equation}\label{eqn:select}
R_p = \tau \times R,
\quad
R_r = (1 - \tau) \times R,
\end{equation}
where $R_p$ is the number of predictor-selected samples and $R_r$ is the number randomly sampled.
This mixed-selection strategy balances exploitation (via the predictor) and exploration (via randomness),
enhancing robustness under uncertain or less-accurate model predictions.

\begin{table*}
\vspace{-0.2in}
\caption{Detailed setting of Engineering Tasks.}
\label{tab:exptask}
\begin{center}
 \begin{threeparttable}
     \resizebox{1.0\linewidth}{!}{
\begin{tabular}{lccccc}
\toprule
Problem    &$\mathbf{x}$&Design Space Dimension &$\mathbf{S}(x)$&Objectives1 & Objectives2 \\ \midrule
DualFSS    &
12*12*2  &
$10^{86}$&
S-Parameters S2,1 &
\refinedv1{$\max_{\mathbf{x}}\min_{u\in [31.5,34.5]}-S(\mathbf{x})(u) $}   &
\refinedv1{$\max_{\mathbf{x}}\max_{u\in [10.5,15.5]}-S(\mathbf{x})(u) $}
  \\

HGA     &
15*20      &
$10^{90}$&
Realized Gain&
\refinedv1{$\max_{\mathbf{x}}\min_{u\in [2.45,2.55]}S(\mathbf{x})(u)$}&  \refinedv1{$\max_{\mathbf{x}}\min_{u\in [5,6]}S(\mathbf{x})(u)$} \\

\bottomrule
\end{tabular}
}

 \end{threeparttable}
 \end{center}
\vspace{-0.15in}
\end{table*}

\section{Experiments}

We conducted experiments on real-world optimization tasks. The experiments answers two key questions: 1) How does \methodname compare to state-of-the-art approaches in terms of performance and robustness? 2) How do the individual components of \methodname contribute to its overall performance?

\textbf{Task Settings.}
Our method is applied  to two real-world engineering tasks: 1) Dual-layer Frequency Selective Surface (DualFSS), used for electromagnetic noise shielding around chips, and 2) High-gain Antenna (HGA),  used in WiFi routers, both involve two optimization objectives. Details of these tasks are provided in Table \ref{tab:exptask} and Appendix.~\ref{supp:sec:setting-details}.

\textbf{Comparison Methods.}

To highlight the strengths and weaknesses of our approach, we compare it against a diverse set of \emph{predictor-based}, \emph{generative}, and \emph{random sampling} methods, which are commonly used
in EMS or antenna design optimization.
The predictor-based methods employ a learned surrogate model (or predictor) to approximate the expensive
simulation function and guide the optimization. We evaluate:
1) Surrogate-assisted Genetic Algorithm (Surrogate-GA) \citep{zhu2020multiplexing};
2) Surrogate-assisted Random Search (Surrogate-RS);
3) Surrogate-assisted Grey Wolf \citep{dong2020surrogate};
4) Surrogate-assisted Gradient Ascent (InvGrad) \citep{trabucco2022design}, which utilizes a predictor to acquire the gradient; \refine{5) Two-stage Data-driven Evolutionary Optimization (TS-DDEO) \citep{zheng2023two} and 6) Surrogate-assisted hybrid swarm optimization algorithm (SAHSO) \citep{li2022surrogate}, which represent recent state-of-the-art developments in surrogate-assisted evolutionary algorithms.}
The generative methods learn a direct mapping from target objectives to feasible designs.
In our experiments, to ensure fair comparisons, these methods are also updated
iteratively in an online fashion. We compare:
1) cGAN (Generative Adversarial Network) \citep{an2021multifunctional};
2) cVAE (Conditional Variational Autoencoder) \citep{lin2022machine};
3) IDN (Inverse Design Network) \citep{ma2020inverse}, which directly predict designs to fulfill goals;
\refinedv1{
4) GenCO \citep{ferber2024genco} which leverages a VAE for generation and a gradient ascent for optimization.
}

\textbf{Evaluation Metrics. 1).}
Aggregation Value of Objectives (Agg Obj): To evaluate the search or generation capabilities of different methods, we compare their optimal performance using the $O(S(x))$. For fair comparison, all methods use the same objective function to guide their optimization or generation process.

\textbf{2).} Single Objective Value (Obj1, Obj2): Considering that structures with similar objective function values can still exhibit differences in quality. Therefore, when the compared methods produce optimal results with closely matched objective function values, we continue to compare the merits of individual criterial $S_{k}(\mathbf{x})$.

\textbf{Implementation Details.}

In the predictor-based approach, we employ ResNet50 as the predictor model, initialized with a dataset of 300 samples. The total number of simulation runs is limited to 1000 to maintain computational efficiency for Surrogate-GA and Surrogate-RS. For methods like InvGrad and the generative approaches (cGAN, cVAE, and IDN), which require higher model accuracy,

we use 6800 and 3800 initial samples for DualFSS and HGA, resulting in final sizes of 7000 and 4000, respectively.
More implementation details are in Appendix.~\ref{supp:sec:impl-details}.

\begin{table*}
\vspace{-0.1in}
\caption{Comparisons on Dual-layer Frequency Selective Surface and High-gain Antenna.}
\label{tab_expfss}
\centering
\begin{threeparttable}
\resizebox{0.85\linewidth}{!}{
\begin{tabular}{l | c c c c | c c c c}
\toprule

\multicolumn{1}{c}{} & \multicolumn{4}{c}{Dual-layer Frequency Selective Surface} & \multicolumn{4}{c}{High-gain Antenna}\\
\cmidrule(lr){2-5} \cmidrule(lr){6-9}
 \multicolumn{1}{c}{Method} & Agg Obj $\uparrow$& Obj1 $\uparrow$ & Obj2 $\uparrow$ & \multicolumn{1}{c}{\#Simulations} & Agg Obj $\uparrow$ &Obj1 $\uparrow$ & Obj2 $\uparrow$ & \#Simulations\\
\midrule
RS           & 7.2824  & 7.2824  & \textbf{36.6861} & \textbf{1000}  & 0.6314  & 0.6314  & 0.7196     & \textbf{1000}     \\
Surrogate-RS & 5.8116  & 5.8116  & 30.2771 & \textbf{1000}  & 3.0857  & 3.0857  & 3.1845     & \textbf{1000}     \\
Surrogate-GA & 4.1946  & 4.1946  & 32.1198 & \textbf{1000}  & 1.5802  & 1.5802  & 4.5598     & \textbf{1000}     \\
Surrogate-GW & 3.1871  & 3.1871  & 4.1321 & \textbf{1000}  & -1.6613  & -0.2088 & -1.6613     & \textbf{1000}     \\
TS-DDEO & 5.5627  & 5.5627  & 9.0000 & \textbf{1000}  & 0.5201  & 1.0441 & 0.5201     & \textbf{1000}     \\
SAHSO & 4.2066  & 4.2066  & 8.0576 & \textbf{1000}  & 0.0589  & 0.0589 & 2.2410     & \textbf{1000}     \\
\midrule
cGAN         & 3.133 & 8.3658 & 3.133 & 7000           & -1.0918 & -1.0918 & -0.6831    & 4000              \\
cVAE         & 8.9294  & 9.5428 & 8.9294 & 7000           & -1.3657 & -1.3657 & -0.5326     & 4000              \\
IDN          & 4.7335  & 4.7335  & 28.8207 & 7000           & -11.9641 & -11.9641  & -7.1459   & 4000              \\
InvGrad      & 2.8941  & 2.8941  & 24.6731 & 7000           & 3.1783  & 3.1783  & 4.4287     & 4000              \\
GenCO      & 1.1819  & 3.9466  & 1.1819 & 7000           & -5.3032  & -5.3032  & 0.6394     & 4000              \\

\rowcolor{pink!30}[1.2ex][1.4ex] \methodname (Ours)  &  \textbf{15.1964}    &  \textbf{15.1964}   &
 31.0443      & \textbf{1000} &  \textbf{3.6595} & \textbf{3.6595}   &  \textbf{6.4820}       & \textbf{1000}\\

\bottomrule
\end{tabular}
}
\end{threeparttable}
\vspace{-0.15in}
\end{table*}

\subsection{Comparisons on Real-World Tasks}
\label{sec:Comparisons_FSS_HGA}
\begin{figure}
  \centering
  \includegraphics[width=0.7\linewidth]
  {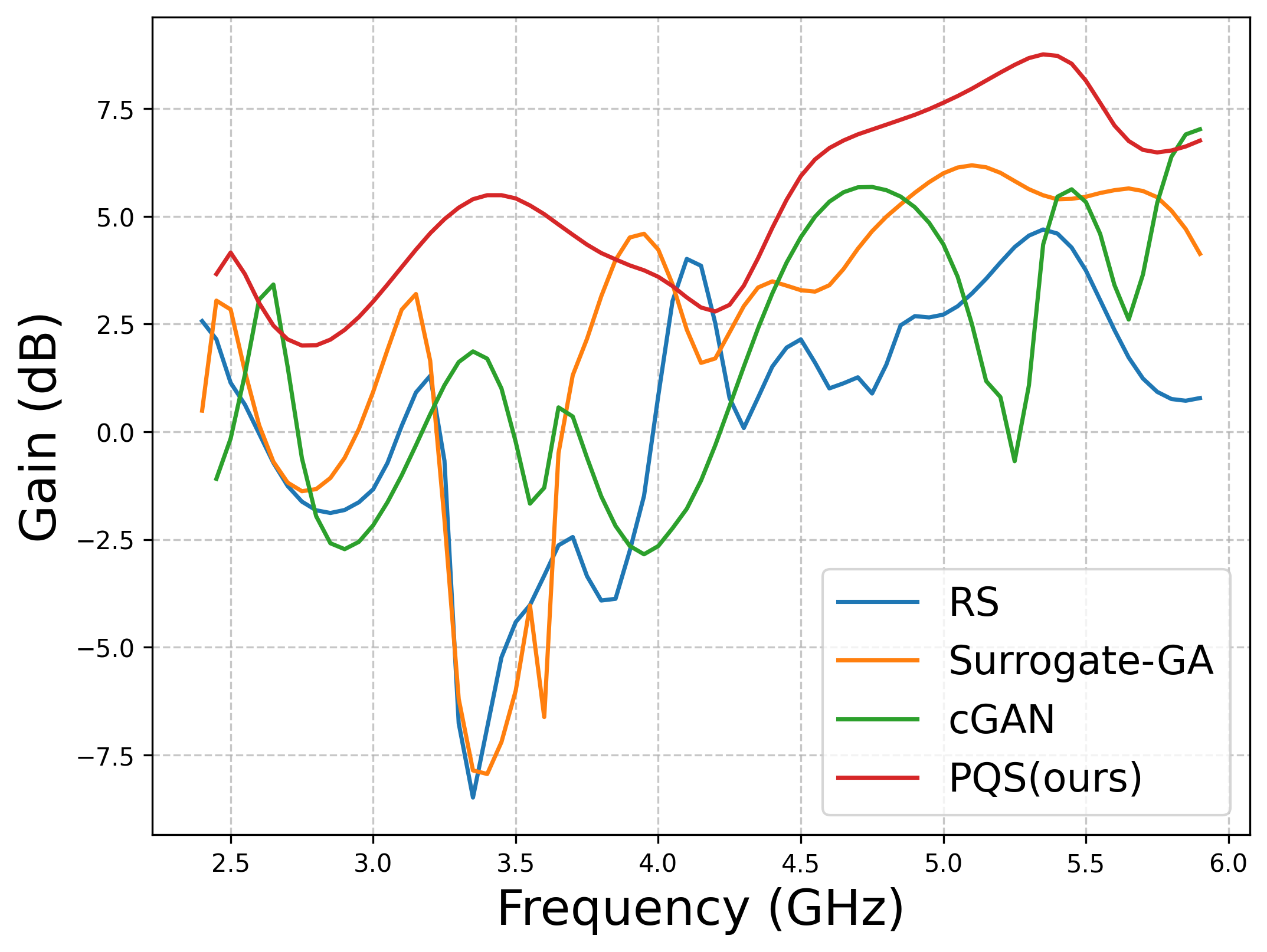}
  \caption{\refinedv1{Simulated Results for High-Gain Antenna Designs obtained from Different Methods}}
  \label{highgain_baseline}
  \vspace{-0.2in}

\end{figure}

Table \ref{tab_expfss} summarizes each method’s performance on two real-world tasks. Additionally, Figure \ref{highgain_baseline} presents a comparison of the performance curves across four experimental groups in the HGA task. We highlight three main observations.

\textbf{\methodname Achieves Both High Aggregate and Balanced Single-Objective Results.} Even with a limited budget of only 1000 simulations, our method attains an aggregate objective of 15.20 dB for DualFSS and 3.66 dB for HGA, markedly surpassing all baselines. Notably, it more than doubles the value of Random Sampling (7.28 dB) under the same budget on DualFSS (a 109\% improvement), while also delivering strong individual objectives (15.20 dB for Obj1 and 31.04 dB for Obj2). A similar trend is observed in HGA, where the improvement is more pronounced. This indicates that \methodname's superiority in overall design quality.

\textbf{Predictor-Based Methods Falter in Few-Shot Settings.} Despite integrating learned predictors, methods such as Surrogate-RS exhibit lower aggregate objective scores (5.81 dB) than even Random Sampling (7.28 dB) on DualFSS, with Surrogate-GW also underperforming in the HGA (-1.66 dB vs 0.63 dB). This suggests that, under a tight simulation budget (1000 samples), the predictors do not receive enough high-quality training data to reliably guide the search. Consequently, the search tends to become trapped in suboptimal regions, highlighting the vulnerability of purely predictor-driven strategies in data-scarce scenarios.

\textbf{Generative Methods Require Larger Budgets but Underperform.} Generative models (e.g., cGAN, cVAE, IDN, GenCO) can be powerful in high-data regimes; however, in this study, we use 7000 simulations, substantially more than the simulations employed by \methodname, yet achieve lower aggregate objectives. This shortfall underscores the advantage of our progressive quadtree representation and consistency-driven mechanism in effectively allocating a limited budget toward maximizing multi-objective outcomes.

In short, \methodname not only outperforms all baselines in the few-shot regime but also demonstrates a robust balance across multiple objectives. These results confirm that our hierarchical representation and sampling strategy can substantially mitigate the data requirements typically faced by predictor- or generator-based solutions.

\begin{table}[t]
  \vspace{-0.1in}
\label{tab:highgain_robustness}
  \centering
  \caption{Robustness Comparisons on High-gain Antenna.}

  \resizebox{1\linewidth}{!}{
\begin{tabular}{l!{\color{black}\vrule}ccc}
    \toprule
    Method
      & Agg Obj $\uparrow$
      & Obj1 $\uparrow$
      & Ob2 $\uparrow$ \\
    \midrule
     Surrogate-RS
      & -1.14 $\pm$ 0.82
      & -0.85 $\pm$ 1.12
      & -0.72 $\pm$ 0.76 \\
     Surrogate-GA
      & -0.93 $\pm$ 1.16
      & -0.65 $\pm$ 1.4
      & 0.36 $\pm$ 1.12 \\
    Surrogate-GW
      & -0.96 $\pm$ 0.48
      & 0.75 $\pm$ 0.92
      & -0.71 $\pm$ 0.86 \\
    TS-DDEO
      & 0.03 $\pm$ 0.42
      & 0.62 $\pm$ 0.87
      & 0.27 $\pm$ 0.36 \\
    SAHSO
      & -0.29 $\pm$ 0.40
      & 0.20 $\pm$ 0.75
      & 0.06 $\pm$ 0.61 \\
      \midrule
    cGAN
      & -2.46 $\pm$ 1.44
      & -2.14 $\pm$ 1.71
      & -0.43 $\pm$ 1.62 \\
    cVAE
      & -4.04 $\pm$ 1.09
      & -3.97 $\pm$ 1.07
      & -1.89 $\pm$ 2.02\\
    IDN
      & -17.08 $\pm$ 4.23
      & -17.08 $\pm$ 4.23
      & -8.08 $\pm$ 3.47 \\
    InvGrad
      & -3.73 $\pm$ 1.72
      & -3.62 $\pm$ 1.91
      & -2.11 $\pm$ 1.48 \\
    GenCO
      & -3.57 $\pm$ 1.38
      & -3.02 $\pm$ 1.96
      & -0.34 $\pm$ 1.58 \\
\rowcolor{pink!30}[1.2ex][1.4ex] \methodname (Ours)
      & \textbf{4.34 $\pm$ 0.34}
      & \textbf{4.45 $\pm$ 0.30}
      & \textbf{4.53 $\pm$ 0.46} \\
    \bottomrule
    \end{tabular}
  }
    \vspace{-0.2in}
\end{table}

\begin{table*}[t]
\begin{minipage}[t]{0.57\linewidth}

\caption{Effectiveness of Number of Variables $N_{max}$.}
\label{tab:aba_n}
\begin{center}
\resizebox{0.78\linewidth}{!}{
\begin{tabular}{l|ccc|c}
\toprule
$N$ &  Agg Obj$\uparrow$ & Obj1(dB)$\uparrow$ & Obj2(dB)$\uparrow$ & Kendall's Tau$\uparrow$ \\
\midrule
16   & 3.0867 & 3.0867 & 6.0295 &
\textbf{0.2838 $\pm$ 0.0527}\\
32  & \textbf{3.6595} & \textbf{3.6595} & \textbf{6.4820} & 0.2324 $\pm$ 0.0523 \\
64  & 3.0233 & 3.0233 & 5.1847 & 0.1255 $\pm$ 0.0297 \\
\bottomrule
\end{tabular}
}
\end{center}

\end{minipage}
\hfill
\begin{minipage}[t]{0.48\linewidth}

\caption{Effectiveness of \modulenameone and \modulenametwo.}
\label{tab:aba_pg}
\begin{center}
     \resizebox{0.78\linewidth}{!}{
\begin{tabular}{cc|ccc}
\toprule
\modulenameone & \modulenametwo & Agg Obj$\uparrow$ & Obj1(dB)$\uparrow$ & Obj2(dB)$\uparrow$ \\
\midrule
  &    & 3.0857 & 3.0857 & 3.1845 \\
\cmark &   & 3.2209 & 3.2209 & 6.3369 \\
\cmark & \cmark   & \textbf{3.6595}& \textbf{3.6595}& \textbf{6.4820} \\
\bottomrule
\end{tabular}
}
\end{center}
\end{minipage}

\end{table*}

\subsection{Robustness Comparisons on High-gain Antenna}
To assess each algorithm’s sensitivity to initialization and data variability, we conducted
10 independent runs for both our PQS method and all baselines.
Because high-fidelity simulations are computationally expensive, we replaced the simulator
with a high-confidence surrogate trained on 37,354 samples from the High-gain
Antenna dataset, achieving a KTau of 0.7969 with respect to real simulations. This ranking correlation is generally sufficient to capture relative performance trends. After fixing the surrogate model’s parameters, we applied each method under identical
configurations. Table~\ref{tab:highgain_robustness}
reports the mean and standard deviation of the aggregate and individual objectives
over all runs. PQS achieves the highest mean values (4.34, 4.45, and 4.53 for Agg Obj, Obj1, and Obj2, respectively)
along with the lowest variance across all objectives, outperforming every baseline
in terms of both peak performance and consistency.

From an engineering perspective, these results suggest that even under challenging,
data-scarce conditions, PQS consistently discovers effective designs, reducing the
risk of suboptimal outcomes due to random initialization or uneven training data. By
contrast, other baselines (for example, Surrogate-RS, cGAN, or InvGrad) show higher
standard deviations, implying increased sensitivity to training randomness and model
inaccuracies. Overall, the combination of high average performance and low variance
reinforces PQS’s robustness and reliability in real-world antenna design, where
resource-intensive simulations demand stable yet efficient optimization strategies.
\subsection{Ablation Studies}
In this section, we dissect our method’s key design choices to understand their individual
contributions and overall impact on performance. Specifically, we investigate:
(1) the effect of varying the maximum dimensionality \( N_{\max} \) in our progressive representation,
(2) the benefits of the quadtree-based search strategy, and
(3) the role of consistency-based sample selection.
All ablation experiments are conducted on the High-gain Antenna (HGA) task for consistency.
utilizing identical training and evaluation settings.

\textbf{Study on the  Number of Variables}.
As outlined in Section \ref{sec:Progressive}, the variable $N_{max}$ shapes the complexity of the design space and directly influences the challenge of identifying high-quality samples. In this section, we examine the impact of varying $N_{max}$ in the context of the HGA task detailed in the Implementation Details section. We evaluate $N_{max}$ values of {16, 32, 64}, and as shown in Table \ref{tab:aba_n}, our \methodname achieves satisfactory performance with an objective function value of 3.66 dB when $N_{max} = 32$. It indicates that a smaller $N_{max}$ is prone to trapping the search in locally sub-optimal regions, while a larger $N_{max}$ makes it difficult to construct accurate models given a limited budget.

We computed the mean and variance of Kendall’s Tau (KTau) for predictors trained on samples generated with varying $N_{max}$ values, with a fixed total sample size of 1000. Samples were split into validation, test, and training sets, and hyperparameter tuning was performed separately for each dataset. Using optimal parameters, we conducted 10 trials per predictor and calculated KTau. Results showed that $N_{max}$ significantly affects KTau, with $N_{max} = 16$ yielding the best performance. As $N_{max}$ increased, KTau decreased, indicating that predictors map the design space more accurately when it is smaller. This supports \modulenameone, showing that expanding from low to high-dimensional spaces enhances predictor performance with limited samples.

\textbf{Effectiveness of Quadtree-based Search Strategy}. We conduct experiments to further demonstrate the effectiveness of our progressive space design. Specifically, we compare our methods with a variant which replaces progressive design strategy with random sampling. Our experiments are conducted in High-gain Antenna under the same computational budget and the results are reported in Table \ref{tab:aba_pg}.
From the results, our method outperform the variant without \modulenameone, generating EMS design with higher objective function value (e.g., 3.22$dB$ vs 3.09$dB$). These results demonstrate the necessity of the proposed progressive search.

\textbf{Effectiveness of Consistency-based Sample Selection}. To verify the effectiveness of the proposed \modulenametwo strategy, we compare our methods with a variant that simply select the \refine{top-K} structures evaluated by the predictor. Our experiments are conducted in High-gain Antenna and we present the objective value in Table \ref{tab:aba_pg}.It is evident that our method outperforms the variant without \modulenametwo, yielding EMS design with higher objective function values (e.g., 3.22 $dB$ vs. 3.09 $dB$). These results show that our method could alleviate the bias induced by inaccurate predictions.

\begin{figure}
  \centering
  \includegraphics[width=0.78\linewidth]
  {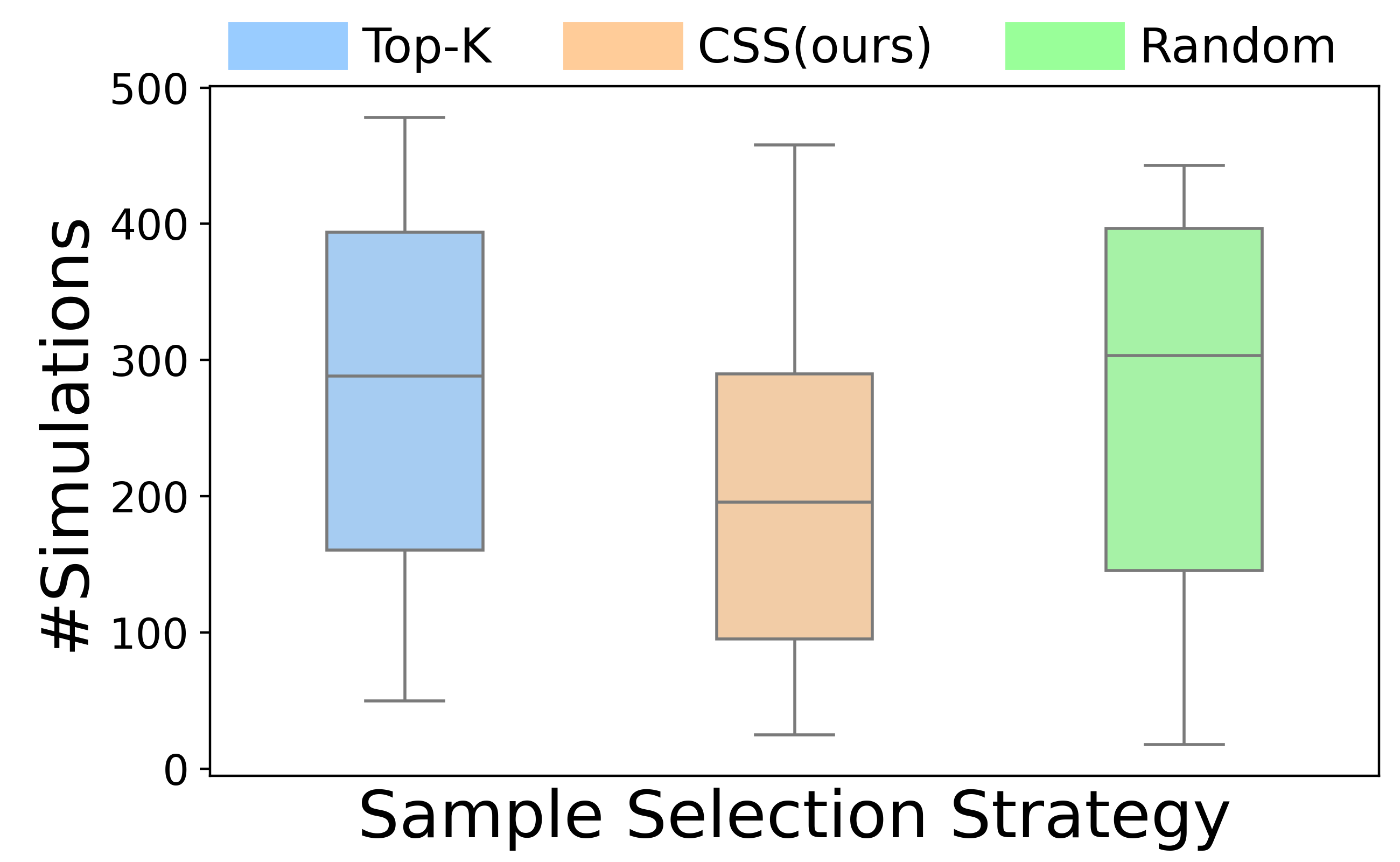}
  \caption{Simulation Costs for Optimal Solution Across Sample Selection Strategy.}
  \label{aba_CSS}

\end{figure}
To more clearly demonstrate the advantages of our method, we designed an additional experiments. A dataset of 1000 samples was randomly divided into two equal parts. The first part was used as the initial training for the predictor.The labels of the second part were masked. We applied three different sample selection strategies—\refinedv1{\modulenametwo} (ours), Top-K, and Random—to identify the actual optimal solution within the set. In each round, 20 samples were selected and added to the training set to update the predictor, with this process continuing until the optimal sample was found. Each experiment was repeated 20 times. The results in Figure \ref{aba_CSS} demonstrate that our method outperforms traditional approaches, improving search efficiency by 50\%.

\section{Conclusion}
In this paper, we propose a Deep Progressive Search method under Limited Data. Specifically, we devise a Quadtree-based Search Strategy
method. By progressively
searching in the simplified space, the quality of samples is
improved, thus reducing dependence on the number of training samples. In addition, we introduce a Consistency-based Sample Selection. With this strategy, the search process can achieve a better balance between exploration and exploitation. Extensive experimental results on
real-world engineering tasks demonstrate the effectiveness of
our method. The results show that our method can achieve satisfactory designs under limited computational budgets and effectively shorten the product designing cycle.
\section*{Acknowledgments}
This work was partially supported by the Joint Funds of the National Natural Science Foundation of China (Grant No.U24A20327).

\section*{Impact Statement}
This paper presents work whose goal is to advance the field of Machine Learning. There are many potential societal consequences of our work, none of which we feel must be specifically highlighted here.
\bibliography{example_paper}

\begin{thebibliography}{40}
\providecommand{\natexlab}[1]{#1}
\providecommand{\url}[1]{\texttt{#1}}
\expandafter\ifx\csname urlstyle\endcsname\relax
  \providecommand{\doi}[1]{doi: #1}\else
  \providecommand{\doi}{doi: \begingroup \urlstyle{rm}\Url}\fi

\bibitem[An et~al.(2021)An, Zheng, Tang, Shalaginov, Zhou, Li, Kang,
  Richardson, Gu, Hu, et~al.]{an2021multifunctional}
An, S., Zheng, B., Tang, H., Shalaginov, M.~Y., Zhou, L., Li, H., Kang, M.,
  Richardson, K.~A., Gu, T., Hu, J., et~al.
\newblock Multifunctional metasurface design with a generative adversarial
  network.
\newblock \emph{Advanced Optical Materials}, 9\penalty0 (5):\penalty0 2001433,
  2021.

\bibitem[Bondeson et~al.(2012)Bondeson, Rylander, and
  Ingelstr{\"o}m]{bondeson2012computational}
Bondeson, A., Rylander, T., and Ingelstr{\"o}m, P.
\newblock \emph{Computational electromagnetics}.
\newblock Springer, 2012.

\bibitem[Brookes et~al.(2019)Brookes, Park, and
  Listgarten]{brookes2019conditioning}
Brookes, D., Park, H., and Listgarten, J.
\newblock Conditioning by adaptive sampling for robust design.
\newblock In \emph{International conference on machine learning}, pp.\
  773--782. PMLR, 2019.

\bibitem[Chen et~al.(2023)Chen, Qian, Zhang, Jia, and
  Chen]{chen2023correlating}
Chen, J., Qian, C., Zhang, J., Jia, Y., and Chen, H.
\newblock Correlating metasurface spectra with a generation-elimination
  framework.
\newblock \emph{Nature Communications}, 14\penalty0 (1):\penalty0 4872, 2023.

\bibitem[Chen et~al.(2015)Chen, Montgomery, and
  Boluf{\'e}-R{\"o}hler]{chen2015measuring}
Chen, S., Montgomery, J., and Boluf{\'e}-R{\"o}hler, A.
\newblock Measuring the curse of dimensionality and its effects on particle
  swarm optimization and differential evolution.
\newblock \emph{Applied Intelligence}, 42:\penalty0 514--526, 2015.

\bibitem[Cheng et~al.(2022)Cheng, Lyu, Li, Ye, Hao, and Yan]{cheng2022policy}
Cheng, R., Lyu, X., Li, Y., Ye, J., Hao, J., and Yan, J.
\newblock The policy-gradient placement and generative routing neural networks
  for chip design.
\newblock \emph{Advances in Neural Information Processing Systems},
  35:\penalty0 26350--26362, 2022.

\bibitem[Daulton et~al.(2022)Daulton, Wan, Eriksson, Balandat, Osborne, and
  Bakshy]{daulton2022bayesian}
Daulton, S., Wan, X., Eriksson, D., Balandat, M., Osborne, M.~A., and Bakshy,
  E.
\newblock Bayesian optimization over discrete and mixed spaces via
  probabilistic reparameterization.
\newblock \emph{Advances in Neural Information Processing Systems},
  35:\penalty0 12760--12774, 2022.

\bibitem[Deng et~al.(2021)Deng, Dong, Ren, Khatib, Soltani, Tarokh, Padilla,
  and Malof]{deng2021benchmarking}
Deng, Y., Dong, J., Ren, S., Khatib, O., Soltani, M., Tarokh, V., Padilla, W.,
  and Malof, J.
\newblock Benchmarking data-driven surrogate simulators for artificial
  electromagnetic materials.
\newblock In \emph{Thirty-fifth Conference on Neural Information Processing
  Systems Datasets and Benchmarks Track (Round 2)}, 2021.

\bibitem[Dong \& Dong(2020)Dong and Dong]{dong2020surrogate}
Dong, H. and Dong, Z.
\newblock Surrogate-assisted grey wolf optimization for high-dimensional,
  computationally expensive black-box problems.
\newblock \emph{Swarm and Evolutionary Computation}, 57:\penalty0 100713, 2020.

\bibitem[Dong \& Yang(2020)Dong and Yang]{dong2020bench}
Dong, X. and Yang, Y.
\newblock Nas-bench-201: Extending the scope of reproducible neural
  architecture search.
\newblock \emph{arXiv preprint arXiv:2001.00326}, 2020.

\bibitem[Ferber et~al.(2024)Ferber, Zharmagambetov, Huang, Dilkina, and
  Tian]{ferber2024genco}
Ferber, A.~M., Zharmagambetov, A., Huang, T., Dilkina, B., and Tian, Y.
\newblock Gen{CO}: Generating diverse solutions to design problems with
  combinatorial nature, 2024.
\newblock URL \url{https://openreview.net/forum?id=V7uPprVelO}.

\bibitem[Gao et~al.(2023)Gao, Tan, and Li]{Gao2023Pifold}
Gao, Z., Tan, C., and Li, S.~Z.
\newblock Pifold: Toward effective and efficient protein inverse folding.
\newblock In \emph{The Eleventh International Conference on Learning
  Representations, {ICLR} 2023, Kigali, Rwanda, May 1-5, 2023}. OpenReview.net,
  2023.

\bibitem[Gaulton et~al.(2017)Gaulton, Hersey, Nowotka, Bento, Chambers, Mendez,
  Mutowo, Atkinson, Bellis, Cibri{\'a}n-Uhalte, et~al.]{gaulton2017chembl}
Gaulton, A., Hersey, A., Nowotka, M., Bento, A.~P., Chambers, J., Mendez, D.,
  Mutowo, P., Atkinson, F., Bellis, L.~J., Cibri{\'a}n-Uhalte, E., et~al.
\newblock The chembl database in 2017.
\newblock \emph{Nucleic acids research}, 45\penalty0 (D1):\penalty0 D945--D954,
  2017.

\bibitem[Han et~al.(2019)Han, Zhang, Zhou, and Wang]{han2019progan}
Han, X., Zhang, L., Zhou, K., and Wang, X.
\newblock Progan: Protein solubility generative adversarial nets for data
  augmentation in dnn framework.
\newblock \emph{Computers \& Chemical Engineering}, 131:\penalty0 106533, 2019.

\bibitem[Jia et~al.(2023)Jia, Qian, Fan, Cai, Li, and Chen]{jia2023knowledge}
Jia, Y., Qian, C., Fan, Z., Cai, T., Li, E.-P., and Chen, H.
\newblock A knowledge-inherited learning for intelligent metasurface design and
  assembly.
\newblock \emph{Light: Science \& Applications}, 12\penalty0 (1):\penalty0 82,
  2023.

\bibitem[Jing et~al.(2022)Jing, Wang, Wu, Ren, Xie, Liu, Ye, Li, Fan, and
  Chen]{jing2022neural}
Jing, G., Wang, P., Wu, H., Ren, J., Xie, Z., Liu, J., Ye, H., Li, Y., Fan, D.,
  and Chen, S.
\newblock Neural network-based surrogate model for inverse design of
  metasurfaces.
\newblock \emph{Photonics Research}, 10\penalty0 (6):\penalty0 1462--1471,
  2022.

\bibitem[Jumper et~al.(2021)Jumper, Evans, Pritzel, Green, Figurnov,
  Ronneberger, Tunyasuvunakool, Bates, {\v{Z}}{\'\i}dek, Potapenko,
  et~al.]{jumper2021highly}
Jumper, J., Evans, R., Pritzel, A., Green, T., Figurnov, M., Ronneberger, O.,
  Tunyasuvunakool, K., Bates, R., {\v{Z}}{\'\i}dek, A., Potapenko, A., et~al.
\newblock Highly accurate protein structure prediction with alphafold.
\newblock \emph{Nature}, 596\penalty0 (7873):\penalty0 583--589, 2021.

\bibitem[Kendall(1938)]{kendall1938new}
Kendall, M.~G.
\newblock A new measure of rank correlation.
\newblock \emph{Biometrika}, 30\penalty0 (1-2):\penalty0 81--93, 1938.

\bibitem[Koziel \& Ogurtsov(2014)Koziel and Ogurtsov]{koziel2014antenna}
Koziel, S. and Ogurtsov, S.
\newblock \emph{Antenna design by simulation-driven optimization}.
\newblock Springer, 2014.

\bibitem[Koziel et~al.(2022)Koziel, {\c{C}}al{\i}k, Mahouti, and
  Belen]{koziel2022low-cost}
Koziel, S., {\c{C}}al{\i}k, N., Mahouti, P., and Belen, M.~A.
\newblock Low-cost and highly accurate behavioral modeling of antenna
  structures by means of knowledge-based domain-constrained deep learning
  surrogates.
\newblock \emph{IEEE Transactions on Antennas and Propagation}, 71\penalty0
  (1):\penalty0 105--118, 2022.

\bibitem[Li et~al.(2022)Li, Li, Cai, and Gao]{li2022surrogate}
Li, F., Li, Y., Cai, X., and Gao, L.
\newblock A surrogate-assisted hybrid swarm optimization algorithm for
  high-dimensional computationally expensive problems.
\newblock \emph{Swarm and Evolutionary Computation}, 72:\penalty0 101096, 2022.

\bibitem[Lin et~al.(2022)Lin, Hou, Wang, Tang, Shi, Tian, Xu,
  et~al.]{lin2022machine}
Lin, H., Hou, J., Wang, Y., Tang, R., Shi, X., Tian, Y., Xu, W., et~al.
\newblock Machine-learning-assisted inverse design of scattering enhanced
  metasurface.
\newblock \emph{Optics Express}, 30\penalty0 (2):\penalty0 3076--3088, 2022.

\bibitem[Liu et~al.(2018{\natexlab{a}})Liu, Zoph, Neumann, Shlens, Hua, Li,
  Fei-Fei, Yuille, Huang, and Murphy]{liu2018progressive}
Liu, C., Zoph, B., Neumann, M., Shlens, J., Hua, W., Li, L.-J., Fei-Fei, L.,
  Yuille, A., Huang, J., and Murphy, K.
\newblock Progressive neural architecture search.
\newblock In \emph{Proceedings of the European conference on computer vision
  (ECCV)}, pp.\  19--34, 2018{\natexlab{a}}.

\bibitem[Liu et~al.(2018{\natexlab{b}})Liu, Simonyan, and Yang]{liu2018darts}
Liu, H., Simonyan, K., and Yang, Y.
\newblock Darts: Differentiable architecture search.
\newblock \emph{arXiv preprint arXiv:1806.09055}, 2018{\natexlab{b}}.

\bibitem[Ma et~al.(2020)Ma, Huang, Pu, Xu, Luo, Guo, and Luo]{ma2020inverse}
Ma, J., Huang, Y., Pu, M., Xu, D., Luo, J., Guo, Y., and Luo, X.
\newblock Inverse design of broadband metasurface absorber based on
  convolutional autoencoder network and inverse design network.
\newblock \emph{Journal of Physics D: Applied Physics}, 53\penalty0
  (46):\penalty0 464002, 2020.

\bibitem[Madan et~al.(2023)Madan, Rector-Brooks, Korablyov, Bengio, Jain, Nica,
  Bosc, Bengio, and Malkin]{madan2023learning}
Madan, K., Rector-Brooks, J., Korablyov, M., Bengio, E., Jain, M., Nica, A.~C.,
  Bosc, T., Bengio, Y., and Malkin, N.
\newblock Learning gflownets from partial episodes for improved convergence and
  stability.
\newblock In \emph{International Conference on Machine Learning}, pp.\
  23467--23483. PMLR, 2023.

\bibitem[Magar et~al.(2022)Magar, Wang, Lorsung, Liang, Ramasubramanian, Li,
  and Farimani]{magar2022auglichem}
Magar, R., Wang, Y., Lorsung, C., Liang, C., Ramasubramanian, H., Li, P., and
  Farimani, A.~B.
\newblock Auglichem: data augmentation library of chemical structures for
  machine learning.
\newblock \emph{Machine Learning: Science and Technology}, 3\penalty0
  (4):\penalty0 045015, 2022.

\bibitem[Majorel et~al.(2022)Majorel, Girard, Arbouet, Muskens, and
  Wiecha]{majorel2022deep}
Majorel, C., Girard, C., Arbouet, A., Muskens, O.~L., and Wiecha, P.~R.
\newblock Deep learning enabled strategies for modeling of complex aperiodic
  plasmonic metasurfaces of arbitrary size.
\newblock \emph{ACS photonics}, 9\penalty0 (2):\penalty0 575--585, 2022.

\bibitem[Naseri et~al.(2022)Naseri, Goussetis, Fonseca, and
  Hum]{naseri2022synthesis}
Naseri, P., Goussetis, G., Fonseca, N.~J., and Hum, S.~V.
\newblock Synthesis of multi-band reflective polarizing metasurfaces using a
  generative adversarial network.
\newblock \emph{Scientific Reports}, 12\penalty0 (1):\penalty0 17006, 2022.

\bibitem[Peurifoy et~al.(2018)Peurifoy, Shen, Jing, Yang, Cano-Renteria,
  DeLacy, Joannopoulos, Tegmark, and
  Solja{\v{c}}i{\'c}]{peurifoy2018nanophotonic}
Peurifoy, J., Shen, Y., Jing, L., Yang, Y., Cano-Renteria, F., DeLacy, B.~G.,
  Joannopoulos, J.~D., Tegmark, M., and Solja{\v{c}}i{\'c}, M.
\newblock Nanophotonic particle simulation and inverse design using artificial
  neural networks.
\newblock \emph{Science advances}, 4\penalty0 (6):\penalty0 eaar4206, 2018.

\bibitem[Ren et~al.(2021)Ren, Xiao, Chang, Huang, Li, Chen, and
  Wang]{ren2021comprehensive}
Ren, P., Xiao, Y., Chang, X., Huang, P.-Y., Li, Z., Chen, X., and Wang, X.
\newblock A comprehensive survey of neural architecture search: Challenges and
  solutions.
\newblock \emph{ACM Computing Surveys (CSUR)}, 54\penalty0 (4):\penalty0 1--34,
  2021.

\bibitem[Shahane et~al.(2023)Shahane, Manjiri, Jain, and
  Kumar]{shahane2023graph}
Shahane, A.~H., Manjiri, S. V.~S., Jain, A., and Kumar, S.
\newblock Graph of circuits with gnn for exploring the optimal design space.
\newblock In \emph{Thirty-seventh Conference on Neural Information Processing
  Systems}, 2023.

\bibitem[Siems et~al.(2020)Siems, Zimmer, Zela, Lukasik, Keuper, and
  Hutter]{siems2020bench}
Siems, J., Zimmer, L., Zela, A., Lukasik, J., Keuper, M., and Hutter, F.
\newblock Nas-bench-301 and the case for surrogate benchmarks for neural
  architecture search.
\newblock \emph{arXiv preprint arXiv:2008.09777}, 4:\penalty0 14, 2020.

\bibitem[Trabucco et~al.(2022)Trabucco, Geng, Kumar, and
  Levine]{trabucco2022design}
Trabucco, B., Geng, X., Kumar, A., and Levine, S.
\newblock Design-bench: Benchmarks for data-driven offline model-based
  optimization.
\newblock In \emph{International Conference on Machine Learning}, pp.\
  21658--21676. PMLR, 2022.

\bibitem[Wang et~al.(2023)Wang, Yang, Hu, Hossain, Liu, Ou, Ye, and
  Wu]{wang2023end}
Wang, Y., Yang, Z., Hu, P., Hossain, S., Liu, Z., Ou, T.-H., Ye, J., and Wu, W.
\newblock End-to-end diverse metasurface design and evaluation using an
  invertible neural network.
\newblock \emph{Nanomaterials}, 13\penalty0 (18):\penalty0 2561, 2023.

\bibitem[Ying et~al.(2019)Ying, Klein, Christiansen, Real, Murphy, and
  Hutter]{ying2019bench}
Ying, C., Klein, A., Christiansen, E., Real, E., Murphy, K., and Hutter, F.
\newblock Nas-bench-101: Towards reproducible neural architecture search.
\newblock In \emph{International conference on machine learning}, pp.\
  7105--7114. PMLR, 2019.

\bibitem[Zheng et~al.(2023{\natexlab{a}})Zheng, Karapiperis, Kumar, and
  Kochmann]{zheng2023unifying}
Zheng, L., Karapiperis, K., Kumar, S., and Kochmann, D.~M.
\newblock Unifying the design space and optimizing linear and nonlinear truss
  metamaterials by generative modeling.
\newblock \emph{Nature Communications}, 14\penalty0 (1):\penalty0 7563,
  2023{\natexlab{a}}.

\bibitem[Zheng et~al.(2023{\natexlab{b}})Zheng, Shi, and Yang]{zheng2023two}
Zheng, L., Shi, J., and Yang, Y.
\newblock A two-stage surrogate-assisted meta-heuristic algorithm for
  high-dimensional expensive problems.
\newblock \emph{Soft Computing}, 27\penalty0 (10):\penalty0 6465--6486,
  2023{\natexlab{b}}.

\bibitem[Zhu et~al.(2022)Zhu, Li, Wei, and Yin]{zhu2022adversarial}
Zhu, E., Li, E., Wei, Z., and Yin, W.-Y.
\newblock Adversarial-network regularized inverse design of frequency-selective
  surface with frequency-temporal deep learning.
\newblock \emph{IEEE Transactions on Antennas and Propagation}, 70\penalty0
  (10):\penalty0 9460--9469, 2022.

\bibitem[Zhu et~al.(2020)Zhu, Qiu, Wang, Sui, Li, Feng, Ma, and
  Qu]{zhu2020multiplexing}
Zhu, R., Qiu, T., Wang, J., Sui, S., Li, Y., Feng, M., Ma, H., and Qu, S.
\newblock Multiplexing the aperture of a metasurface: inverse design via
  deep-learning-forward genetic algorithm.
\newblock \emph{Journal of Physics D: Applied Physics}, 53\penalty0
  (45):\penalty0 455002, 2020.

\end{thebibliography}
\bibliographystyle{icml2025}

\newpage
\appendix
\onecolumn

\section{Supplementary Materials}

In the supplementary, we provide more implementation details and more experimental results of our \methodname.
We organize our supplementary as follows.

\begin{itemize}[left=0em]

    \item
    In Section ~\ref{supp:sec:setting-details}, we provide more details of the considered two real-world challenging electromagnetic structures, dual-layer frequency selective surface and high-gain antenna

    \item
    In Section~\ref{supp:sec:impl-details}, we depict more implementation details of our \methodname and the compared methods.

    \item In Section~\ref{supp:sec:more-restuls}, we give more experimental results to demonstrate the effectiveness of our \methodname.

\end{itemize}

\section{More Details on Electromagnetic Structures}\label{supp:sec:setting-details}

\textbf{Dual-layer Frequency Selective Surface}.
The dual-layer Frequency Selective Surface (DualFSS) is an electromagnetic structure specifically designed for selectively filtering electromagnetic waves. It consists of two layers of conductive elements, each containing a grid or array of metallic elements exhibiting specific resonant behavior at particular frequencies. This configuration endows the DualFSS with frequency-dependent transmission and reflection characteristics. Despite its more intricate structure compared to a single-layer FSS, the DualFSS provides higher degrees of freedom, allowing for more flexible performance adjustments across different frequencies. In practical applications, the DualFSS finds common usage in scenarios demanding enhanced performance and broader frequency coverage, such as RF communication, radar systems, and engineering designs within the radio frequency spectrum. In our experiments, we utilized the proposed methodology to design the structure of the DualFSS, focusing on the two layers of metallic grids. The designed structure aims to meet specific performance criteria, with detailed parameters and expected metrics outlined below.

\begin{figure}[h]
  \vspace{-0.0 in}
  \setlength{\abovecaptionskip}{-1pt}
  \centering
  \includegraphics[width=0.6\linewidth]{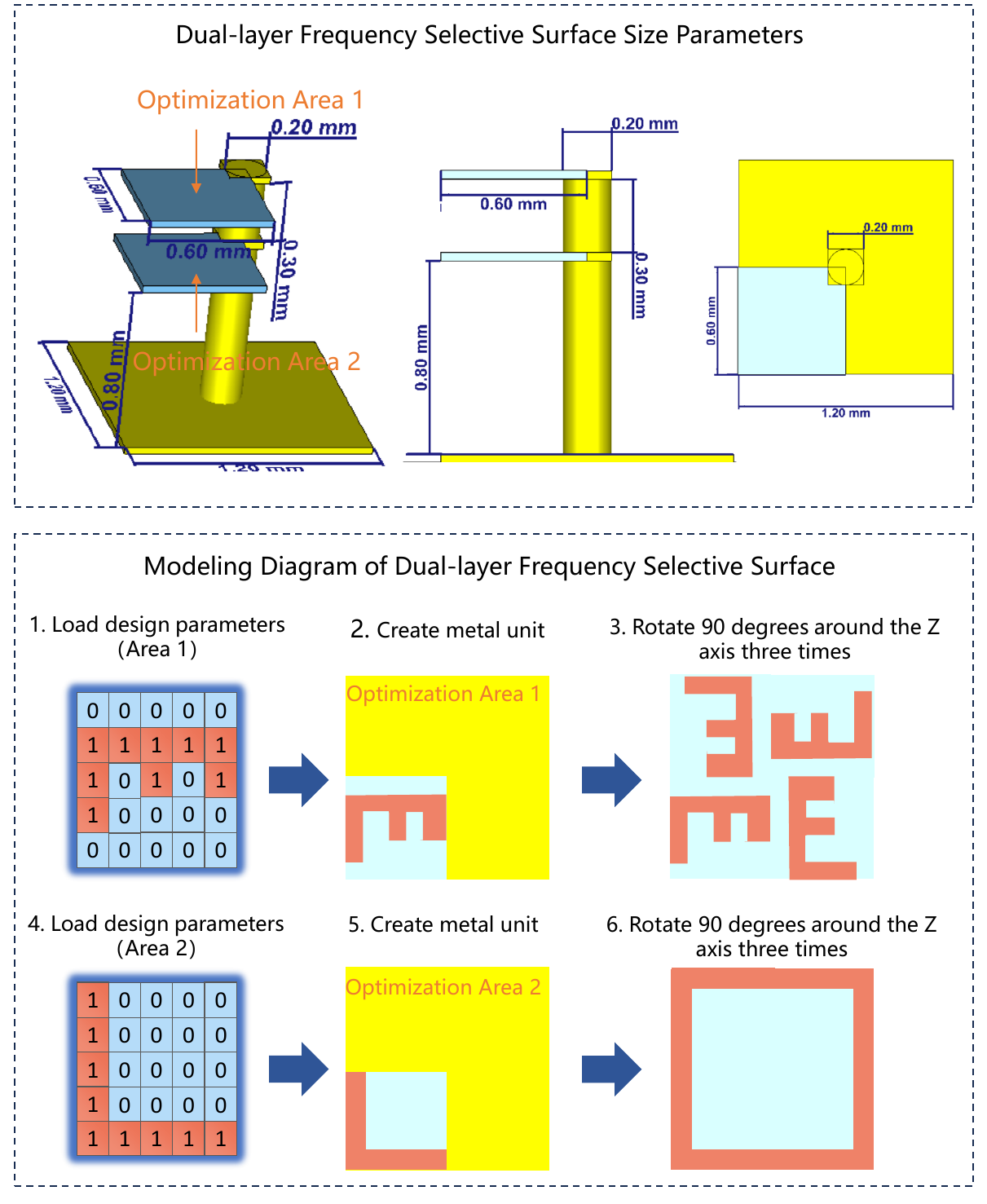}
  \caption{The detailed settings of the Dual-layer Frequency Selective Surface.}
  \label{fig:FSS}
  \vspace{-0.2in}
\end{figure}

Specifically, the DualFSS under investigation is depicted in Figure~\ref{fig:FSS}. The FSS is composed of three layers of boards, each having a thickness of 0.035mm, and features cylindrical elements with a radius of 0.1mm. It is noteworthy that the bottommost layer, representing the grounded metallic surface, remains unaltered throughout the optimization process. In contrast, the upper two layers, initially comprising entirely of air, constitute the optimization space.
The objective of the optimization is to strategically convert certain regions within the air layers into metallic elements, adhering to the constraint that each designed metallic block must have a minimum size of 0.2*0.2mm. The optimization encompasses determining the specific configuration of metallic blocks within the upper layers to achieve desired electromagnetic properties. Importantly, the maximum extent of the air region available for optimization corresponds to the footprint of the bottommost layer.

In terms of optimization objectives, this scenario aims to eliminate electromagnetic noise caused by ultra-high-frequency circuits, preventing such noise from interfering with the operation of mobile phone cameras. The high-frequency noise primarily occurs in two frequency bands: 10.5–15.5 GHz and 31.5–34.5 GHz. We evaluate the suppression capability using the S-Parameter S2,1, which is a parameter that takes only negative values. A smaller magnitude of this parameter indicates stronger suppression performance. We aim to achieve broad absorption capability in the 31.5–34.5 GHz range, while maximizing suppression in the 10.5–15.5 GHz range. To enhance generalization, it is necessary to minimize the maximum value within the 31.5–34.5 GHz band, while minimizing the minimum value within the 10.5–15.5 GHz band to strengthen absorption performance. Accordingly, we define the first objective as minimizing the maximum value of S-Parameters S2,1 over the 31.5–34.5 GHz band, represented by the formula \refinedv1{$\min_{\mathbf{x}}\max_{u \in [31.5, 34.5]} S(\mathbf{x})(u)$}. By taking the inverse, we can transform it into a maximization problem and define it using a new mathematical expression \refinedv1{$\max_{\mathbf{x}}\min_{u \in [31.5, 34.5]} -S(\mathbf{x})(u)$}, and denote this as Obj1. Similarly, the second objective is to minimize the \refinedv1{minimum} value of S-Parameters S2,1 over the 10.5–15.5 GHz band, defined by the formula \refinedv1{$\min_{\mathbf{x}}\min_{u \in [10.5, 15.5]} S(\mathbf{x})(u)$}. We can also transform it into a maximization problem and represented by the formula \refinedv1{$\max_{\mathbf{x}}\max_{u \in [10.5, 15.5]} -S(\mathbf{x})(u)$}, and referred to as Obj2. In this formulation, $\mathbf{x}$ represents the vector of structural design parameters, while \refinedv1{$u$} denotes the frequency in GHz, serving as the independent variable across both frequency bands. The term \refinedv1{$S(\mathbf{x})(u)$} refers to the S-Parameters S2,1 of the FSS for a given design $\mathbf{x}$ at a specific frequency $t$.

An aggressive objective function is set to maximize the worst-case performance of the single objective to achieve balanced shielding capabilities. This is mathematically expressed as $\max_{\mathbf{x}} \left( \min \left( Obj_1(\mathbf{x}), Obj_2(\mathbf{x}) \right) \right)$, where $Obj_1(\mathbf{x})$ represents \refinedv1{$\min_{u \in [31.5, 34.5]} -S(\mathbf{x})(u)$}  while $Obj_2(\mathbf{x})$ represents \refinedv1{$\max_{u \in [10.5, 15.5]} -S(\mathbf{x})(u)$}.

\textbf{High-gain Antenna}.
The high-gain antenna is a specialized electromagnetic structure designed to achieve significant directional amplification of radio frequency signals. This type of antenna is characterized by its ability to focus transmitted or received signals in a specific direction, resulting in a concentrated radiation pattern. In the design of it, the integration of a metal array plays a crucial role in shaping the antenna's radiation pattern and achieving enhanced performance. The metal array structure involves a carefully arranged grid or array of metallic elements, such as reflectors and directors, strategically positioned to optimize the antenna's gain and directional characteristics. High-gain antennas find extensive applications in scenarios requiring long-range communication, satellite communication, and situations where a concentrated signal strength is essential.

\begin{figure}[t]
  \vspace{-0.0in}
  \setlength{\abovecaptionskip}{-1pt}
  \centering
  \includegraphics[width=0.6\linewidth]{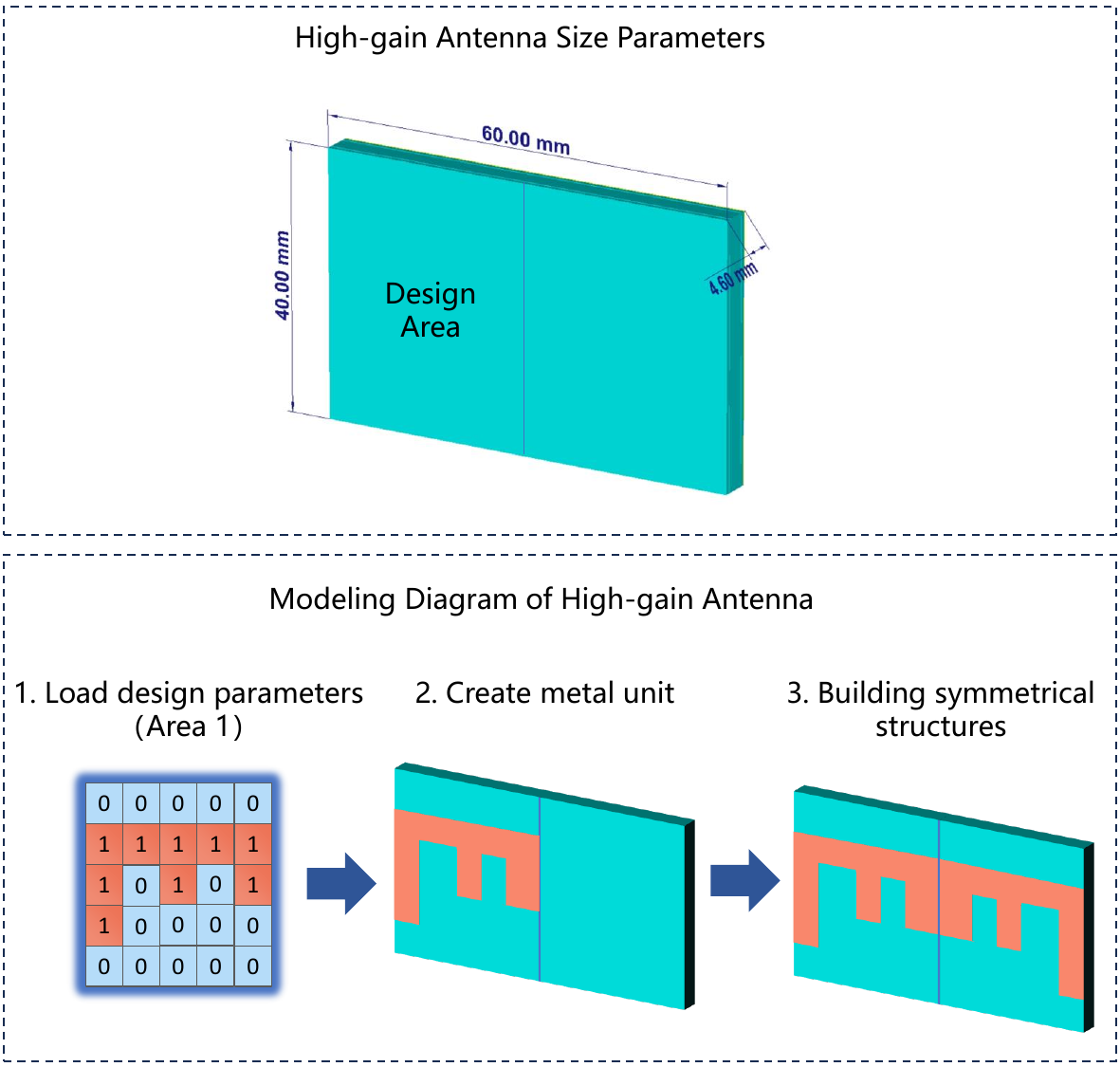}
  \caption{Designs of the High-gain Antenna.}
  \label{fig:hga}
  \vspace{-0.2in}
\end{figure}

Specifically, our structure is a rectangular prism with dimensions of 60 mm in length, 40 mm in width, and 4.6 mm in thickness in Figure~\ref{fig:hga}. The prism is then divided into two halves along the midpoint of its length, parallel to the width. One half is designated as the design region, and after the design is completed, it is mirrored across the vertical plane of symmetry.

The target of this scenario is to design a dual-band router antenna operating at both 2.4 GHz and 5 GHz, ensuring optimal communication performance in both frequency bands simultaneously, rather than having one band perform well while the other lags. Consequently, our two objective functions represent the minimum Realized Gain for the 2.4 GHz and 5 GHz bands, respectively. An Aggregation Value of Objective is set to maximize the worst-case performance across both objectives, aiming to achieve effective dual-band communication.

Specifically, since the minimum value within a given range dictates the weakest communication capability of that band, to meet the communication requirements of the 2.4 GHz band, the first objective is defined as maximizing the minimum Realized Gain in the 2.45–2.55 GHz frequency range. This is mathematically expressed as \refinedv1{$\max_{\mathbf{x}} \min_{u \in [2.45, 2.55]} S(\mathbf{x})(u)$} and denoted as Obj1. In this formulation, $x$ represents the vector of structure design, while \refinedv1{$u$} denotes the frequency in GHz, acting as the independent variable across both frequency bands. The term \refinedv1{$S(\mathbf{x})(u)$} denotes the Realized Gain of the antenna for a given design $x$ at a specific frequency \refinedv1{$u$}.  Similarly, to ensure robust communication in the 5 GHz band, the second objective is defined as maximizing the minimum Realized Gain in the 5.0–6.0 GHz frequency range, expressed as \refinedv1{$\max_{\mathbf{x}} \min_{u \in [5, 6]} S(\mathbf{x})(u)$} and denoted as Obj2.

To achieve strong communication performance across both bands, the smaller of the two objective values is chosen as the Aggregation Value of Objective. This is mathematically expressed as:
\refinedv1{$$\max_{\mathbf{x}} \min \left( \min_{u \in [2.45, 2.55]} S(\mathbf{x})(u), \min_{u \in [5, 6]} S(\mathbf{x})(u) \right).$$} This formulation ensures that the EMS design is optimized for both frequency bands by focusing on improving the worst-case communication performance across the two bands, thereby achieving balanced and robust dual-band communication.

\section{More Implementation Details}\label{supp:sec:impl-details}

\textbf{\methodname (ours)}. For the proposed \methodname, we maintain a consistent setup across both experimental scenarios. In both cases, the initial dataset comprises 300 samples, derived from a progressive design strategy, which allows for more adaptive and efficient sampling.  We utilize ResNet50 as the predictor, ensuring consistency in model structure across different approaches and facilitating a fair comparison of performance. The design variable $N_{max}$ is set to 32. Additionally, when implementing CSS, we estimated the initial $\tau$ (at $t=0$) by training multiple predictors with independent random initializations.
\refinedv1{We set maximum iteration $M = 10000000$ and $K = 10$. The total number of samples $R$ in CSS is $10$.}

\textbf{Compared Methods.} We re-implement the following state-of-the-art electromagnetic structures design methods in our two challenging task, dual-layer frequency selective surface and high-gain antenna. More details of the baseline design methods are provided in the subsequent discussion.

\begin{itemize}[left=0em]

    \item
    \textbf{Random Sampling (RS)}. In this approach, we randomly generate $1000$ electromagnetic structures, evaluating them with the simulation software.

    \item
    \textbf{Bayesian Optimization} \citep{daulton2022bayesian}. After attempting, Bayesian optimization failed to generate non-empty structures, and the updates of new samples did not alter the algorithm's behavior. We believe that Bayesian methods are not effectively applicable to our scenario due to the curse of dimensionality. Therefore, we did not include them in the comparative experiments.

    \item
    \textbf{Surrogate-assisted Random Search (Surrogate-RS)}.

    \refinedv1{This approach adopts the ResNet50 architecture as the surrogate model in both scenarios, consistent with the setup in \methodname, to ensure a fair comparison between the methods.}
    The surrogate model is designed to provide prediction of the objective function given the input of the electromagnetic structures and is ultilized to guide the random search.

    \refinedv1{Specifically, we begin by randomly sampling 300 electromagnetic structures to train an initial surrogate model. In each subsequent iteration, the surrogate model guides the selection of the Top-K samples from M randomly sampled candidates (where $M \gg K$), which are then evaluated through simulation. The surrogate model is updated accordingly, and this process is repeated until the total number of simulated samples reaches the predefined limit of 1000. Finally, we select the best one as optimized result. }
   In practice, we set $M=200000$ and $K=10$ for our experiments in both two real-world tasks.

    \item
    \textbf{Surrogate-GA }\citep{zhu2020multiplexing}. This method exploit a surrogate model to accelerate the evolutionary algorithm. Specifically, this method fit a DNN-based surrogate model with a simulated dataset to assist fitness evaluation during the evolution process. In particular, we have adapted the mutation operator to suit our electromagnetic structural scenario by modifying it to perform transformations between 0 and 1 in the matrix elements. In our experiments, we use the ResNet50 model as the surrogate model. The model’s batch size is set to 256, trained for 200 epochs, with a learning rate of 0.01. First, 300 samples were obtained through random sampling, and simulations were performed on these samples to train the surrogate model. Based on the surrogate model, we set $K$=10 for our experiments, meaning that in each generation, the top 10 samples are selected using a Top-K strategy for simulation verification. The surrogate model is then updated with these results, and a new population is generated. This process continues until the total number of simulated samples reaches 1000.

    \item
    \textbf{Surrogate-GW}\citep{dong2020surrogate}. The authors introduced a method based on the Grey Wolf Optimizer (GWO) for designing electromagnetic structures that satisfy specific electromagnetic performance requirements. GWO is a metaheuristic optimization algorithm inspired by the hunting behavior of grey wolves in nature. By simulating the hierarchical social structure of a wolf pack ($\alpha$, $\beta$, $\delta$, $\omega$) and the dynamic processes of encircling, chasing, and attacking prey, GWO demonstrates excellent global search and local exploitation capabilities in complex high-dimensional optimization problems. In this study, the initial population consists of \refine{500} randomly sampled simulation samples for DualFSS and \refine{300} for HGA. Subsequently, GWO simulates the cooperative hunting mechanism of a wolf pack by continuously adjusting the positions of the leader wolves ($\alpha$, $\beta$, $\delta$) and the remaining wolves, guiding them toward the current optimal individual while maintaining population diversity to prevent premature convergence to local optima. Regarding the experimental setup, the parameter a, which controls the balance between exploration and exploitation during the search process, is set to 3. As the same as Surrogate-GA, we use the ResNet50 model as the surrogate model. The model’s batch size is set to 256, trained for 200 epochs, with a learning rate of 0.01. Based on the surrogate model, we set $K$=10 for our experiments, meaning that in each generation, the top 10 samples are selected using a Top-K strategy for simulation verification. The surrogate model is then updated with these results, and a new population is generated. This process continues until the total number of simulated samples reaches 1000.

    \item
    \textbf{InvGrad} \citep{trabucco2022design}. Trabucco et al. introduces a simple baseline method based on gradient ascent. In this approach, a \refinedv1{ResNet50-based} surrogate model is initially trained on a dataset of electromagnetic structures, which is designed to establish an accurate mapping between the electromagnetic structure and the objective function. \refinedv1{Specifically, the dataset consists of 6800 randomly sampled simulation samples for DualFSS and 3800 for HGA, which represent the minimum number of samples required to ensure the stability and reliability of the model without compromising its performance.} Subsequently, the method performs multiple gradient updates on the input electromagnetic structure based on the surrogate model output, ultimately yielding an satisfactory electromagnetic structure that satisfies the specified criteria. The gradient update could be formulated as $x_{t+1} \gets x_{t} + \alpha\nabla_{x}f(x)$, where $t$ represents the update step and $\alpha$ denotes the learning rate. In pratice, we set $T=1000$, $\alpha=0.01$ for two design tasks. We randomly sampled 10 electromagnetic structures, input them into the surrogate model for optimization,

    \refinedv1{and forwarded the optimized results to the simulation software for evaluation. The surrogate model is then updated with these results, and this process is repeated until the total number of simulated samples reaches the predefined limit of 7000 for DualFSS and 4000 for HGA.}

    \item
    \textbf{IDN} \citep{ma2020inverse}.
    Ma et al. introduces a baseline method for inverse design based on Convolutional Autoencoder Network (CAN) and Inverse Design Network (IDN). In this approach, the authors utilized CAN to compresses input spectrums with the dimension of 1*1000 into low-dimensional spectrums with the dimensionof 1*50. Subsequently, the compressed latent space values were fed into the IDN with the expectation of generating structures that conform to the input spectrum. In our experiments, we set the target values as either the maximum or minimum values within a specific frequency range, eliminating the involvement of high-dimensional inputs. Consequently, we exclusively adopted the IDN component of the method for our purposes.In terms of network architecture, we introduced an additional fully connected layer before the first convolutional layer of the Inverse Design Network. This layer elevates the input target values to a 50-dimensional space to align with subsequent dimensions. The model's batch size is set to 128, trained for 200 epochs, with a learning rate of 0.0001. Adam optimizer is employed, and MAE(Mean Absolute Error) is used for loss computation. \refinedv1{Initially, we performed random sampling to generate 6800 samples for DualFSS and 3800 samples for HGA. These samples were simulated to calculate their respective objective values, which were subsequently used to train the initial models. In the next stage, the models were utilized to generate 10 additional samples, which underwent simulation-based validation. The validated samples were then added to the dataset, and the models were retrained iteratively. This process was repeated until the simulation budget of 7000 and 4000 was reached for DualFSS and HGA, respectively. Ultimately, the sample with the best simulation performance during this process was selected as the final result.}
    \item
    \textbf{cGAN} \citep{an2021multifunctional}
    An et al. presents a generative adversarial network that can generate metasurface designs to meet design goals . Generative adversarial nets can be extended to a conditional model if both the generator and discriminator are conditioned on some extra information. It could be any kind of auxiliary information, such as class labels or data from other modalities. We can perform the conditioning by feeding extra information into the both the discriminator and generator as additional input layer. Consequently, cGAN introduces extra information as conditions in both the encoder and decoder inputs to confer the ability to generate pecific structures based on varying conditions.The model’s batch size is set to 64, trained for 200 epochs, with a discriminator learning rate of 0.00005 and generator learning rate of 0.0002. In addition, the latent dimension is set to 100, Adam optimizer is employed. \refinedv1{We began by randomly sampling 6800 instances for DualFSS and 3800 instances for HGA, followed by simulations to derive their objective values for training the initial models. Using these models, 10 new samples were generated and validated through simulations. These validated samples were incorporated into the dataset, and the models were updated iteratively. This iterative procedure continued until the simulation budgets—7000 for DualFSS and 4000 for HGA—were exhausted. The final result was determined by selecting the sample exhibiting the optimal performance during simulations.}
    \item
    \textbf{cVAE} \citep{lin2022machine}.
    Lin et al. introduces an approach utilizing Conditional Variational Autoencoder (cVAE) to generate metasurface retroreflectors (MRF) structures satisfying specified performance criteria. cVAE represents a variant incorporating both Variational Autoencoder (VAE) and Autoencoder (AE) principles. While VAE extends the encoding-decoding training paradigm of AE by transforming it from encoding inputs into a single point in latent space to encoding inputs into a distribution in latent space, endowing it with generative capabilities, the generated content is inherently uncontrollable. Consequently, cVAE introduces conditions in both the encoder and decoder inputs to confer the ability to generate specific structures based on varying conditions.The model's batch size is set to 128, trained for 200 epochs, with a learning rate of 0.0005. In addition, the latent dimension is set to 20, Adam optimizer is employed, and the loss function is obtained through linear summation of Mean Squared Error (MSE) and 0.00000001 times the Kullback-Leibler (KL) divergence. \refinedv1{An initial random sampling of 6800 samples for DualFSS and 3800 samples for HGA was conducted, with simulations performed to compute the corresponding objective values for initial model training. The trained models then produced 10 new samples, which were subjected to simulation validation. These validated samples were appended to the dataset, and the models were retrained iteratively until the simulation budgets of 7000 for DualFSS and 4000 for HGA were fully utilized. The final output was chosen as the sample demonstrating the highest simulation performance during the process.}
    \item
    \refinedv1{\textbf{GenCO} \citep{ferber2024genco}}
   \refinedv1{
GenCO utilizes VAE (Variational Autoencoders) to generate a variety of designs that account for specific constraints, such as those encountered in nanophotonic materials. Following the approach outlined in the original paper, we integrate electromagnetic structure performance as a constraint objective into the model’s training loss function.
GenCO requires computing the gradient of the objective function with respect to the design variables. However, in our case, obtaining such gradient information directly through simulation is unavailable. To overcome this challenge, we use a surrogate model to approximate the gradient. The surrogate model, implemented using ResNet50, serves to predict the objective function's gradient efficiently. The training parameters of the surrogate model, such as the network architecture and optimization procedure, are consistent with those used in similar works.
Since the original paper does not provide detailed hyperparameter settings for VAE, we made reasonable choices based on standard practices for training generative models. We use a four-layer convolutional and transposed convolutional network architecture for the VAE model. The specific training parameters for our implementation are as follows: latent dimension = 256, number of embeddings = 512, learning rate = 1e-3, and the number of epochs = 100.
}
    \item
    \textbf{TS-DDEO} \citep{zheng2023two}.
    \refine{
    Zhen et al. introduces a two-stage data-driven evolutionary optimization algorithm for solving computationally expensive in high dimensions optimization problems. In the first stage, a surrogate-assisted hierarchical particle swarm optimization method (SHPSO) is used to identify a promising area in the entire search space. In the second stage, a best data-driven optimization method (BDDO) is used to improve the convergence speed in the promising area. In our electromagnetic structure scenario, we used the ResNet50 model as the surrogate model of TS-DDEO and randomly generated and evaluated 500 simulation samples of DualFSS (300 for HGA) to train the initial surrogate model. The batch size of this model was set to 256, and it was trained for 200 epochs, with a learning rate of 0.01. Consistent with the parameters set in the original paper, we set the population size of DE sampling in SHPSO and BDDO to 50, and the probability r of FC strategy in the BDDO phase was 0.2. In addition, out of a total budget of 1000 simulation evaluations, we allocated 300 evaluation budgets for the first phase, and then switched to the second phase. After searching for candidate populations in each stage, we select the top 10 candidate solutions for simulation verification. Then, we use the simulated results to update the historical database and retrain the ResNet50 model.}

    \item
    \textbf{SAHSO} \citep{li2022surrogate}.
    \refine{
    Li et al. propose a surrogate-assisted hybrid swarm optimization algorithm (SAHSO) for high-dimensional computationally expensive problems. In the first stage, the Teaching-learning-based optimization (TLBO) learner phase is employed to enhance global exploration through random interactions among individuals. In the second stage, the algorithm switches to standard particle swarm optimization (SPSO) to accelerate local exploitation, with the transition controlled by $T_0$. In our scenario, we adapted SAHSO with modifications to address high dimensionality. To enhance exploration, we increase the duration of the first stage by setting $T_0$ to 60\% of the maximum evaluation budgets. During the second stage, to ensure both computational efficiency and fair comparison with other methods, the number of pre-screened candidate solutions was limited to 10–15 per iteration. The original surrogate model RBF was replaced with ResNet50. In each iteration, we randomly sample 30,000 candidates, evaluate them using ResNet50, and select the one with the highest predicted performance as the surrogate optimum for updating the global best. The ResNet50 model was trained using 500 initial simulation samples for DualFSS and 300 for HGA, with a batch size of 256, 200 training epochs, and a learning rate of 0.01.}

\end{itemize}

\refine{\textbf{Time-saving Calculation.} We present a detailed breakdown of the average computational time savings achieved by our method compared to generative approaches. Based on our statistics, the average simulation time per sample is 583.7 seconds on Dual-layer Frequency Selective Surface and 558.8 seconds on the High-gain Antenna. The time-saving calculation is as follows:
\begin{itemize}[left=0em]
    \item
    \textbf{Dual-layer Frequency Selective Surface}: Compared to generative approaches, our method reduces the number of required simulations by 3,000. The total time saved is $\frac{583.7 \times 3,000}{3,600 \times 24} \approx 20.27$ days.
    \item
    \textbf{High-gain Antenna}: Compared to generative approaches, our method reduces the number of required simulations by 6,000. The total time saved is $\frac{558.8 \times 6,000}{3,600 \times 24} \approx 38.80$ days.
\end{itemize}
These results clearly demonstrate that our method achieves significant efficiency gains (75$\sim$85\% cost reduction) compared to generative approaches.
}

\section{More Experimental Results}\label{supp:sec:more-restuls}

\begin{table}
\vspace{-0.2in}
\caption{Effect of Importance Assignment.}
\label{tab:aba_ia}
\begin{center}
     \resizebox{0.7\linewidth}{!}{
\begin{tabular}{l|ccc}
\toprule
Method &  Objective Function Value$\uparrow$&Objective1(dB)$\uparrow$&Objective2(dB)$\uparrow$\\
\midrule

\methodname w/o IA & 2.32 & 2.32 & 5.57 \\
\methodname  & \textbf{3.66} & \textbf{3.66} & \textbf{6.48} \\

\bottomrule
\end{tabular}
}
\end{center}
\end{table}

\begin{figure}[htb]
  \centering
    \includegraphics[width=0.35\linewidth]{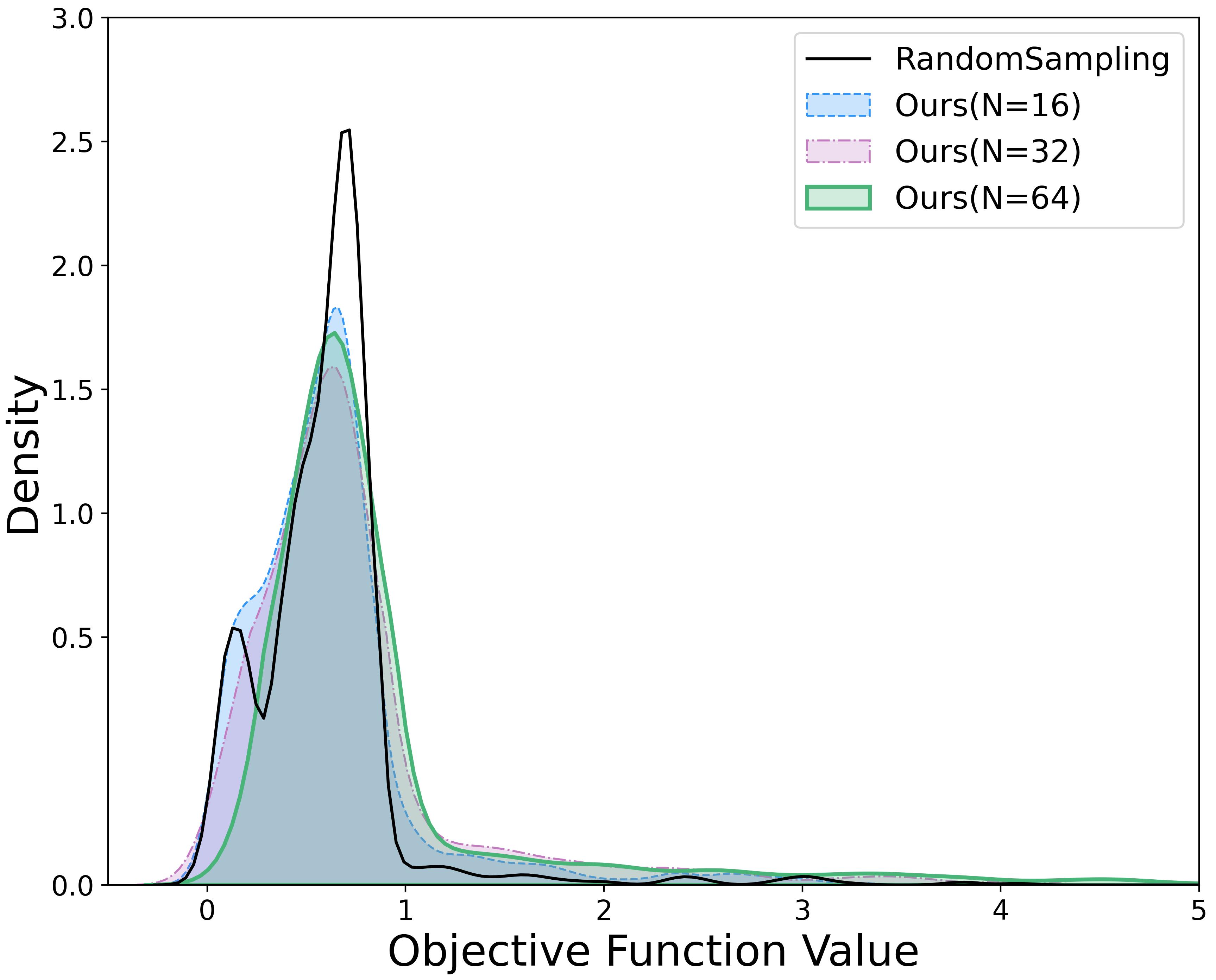}
    \vspace{-0.2in}
  \caption{Comparison of Sample Performance Distribution under Different Parameter $N$ Settings.}
  \label{fig:N_obj}
  \vspace{-3mm}
\end{figure}

\textbf{Effect of the  Number of Variables $N_{max}$}. To further illustrate the effect of the parameter $N_{max}$, we present kernel density estimation (KDE) plots of the sample performance distribution under different parameter settings in Figure \ref{fig:N_obj}. We sampled and simulated 1000 samples for each parameter setting and random sampling. The experimental results demonstrate that our method is more likely to sample higher-performing instances across various parameter configurations, whereas most of the samples generated through random sampling tend to cluster in the lower performance range.

\textbf{Effect of Depth-wise Importance Assignment}. We investigate the effect of the depth-wise importance assignment. For a fair comparison, we conduct this experiment under the same simulation budget in high-gain antenna design task. From Table \ref{tab:aba_ia}, without importance assignment, the \methodname tends to find sub-optimal electromagnetic structure. When equipped with the proposed importance assignment, the searched structure consistently outperforms that without importance assignment (e.g., 3.66$dB$ vs 2.32$dB$). These findings illustrate the essential nature and efficacy of the introduced depth-wise importance assignment.

\begin{figure}
  \centering
  \includegraphics[width=0.9\linewidth]
  {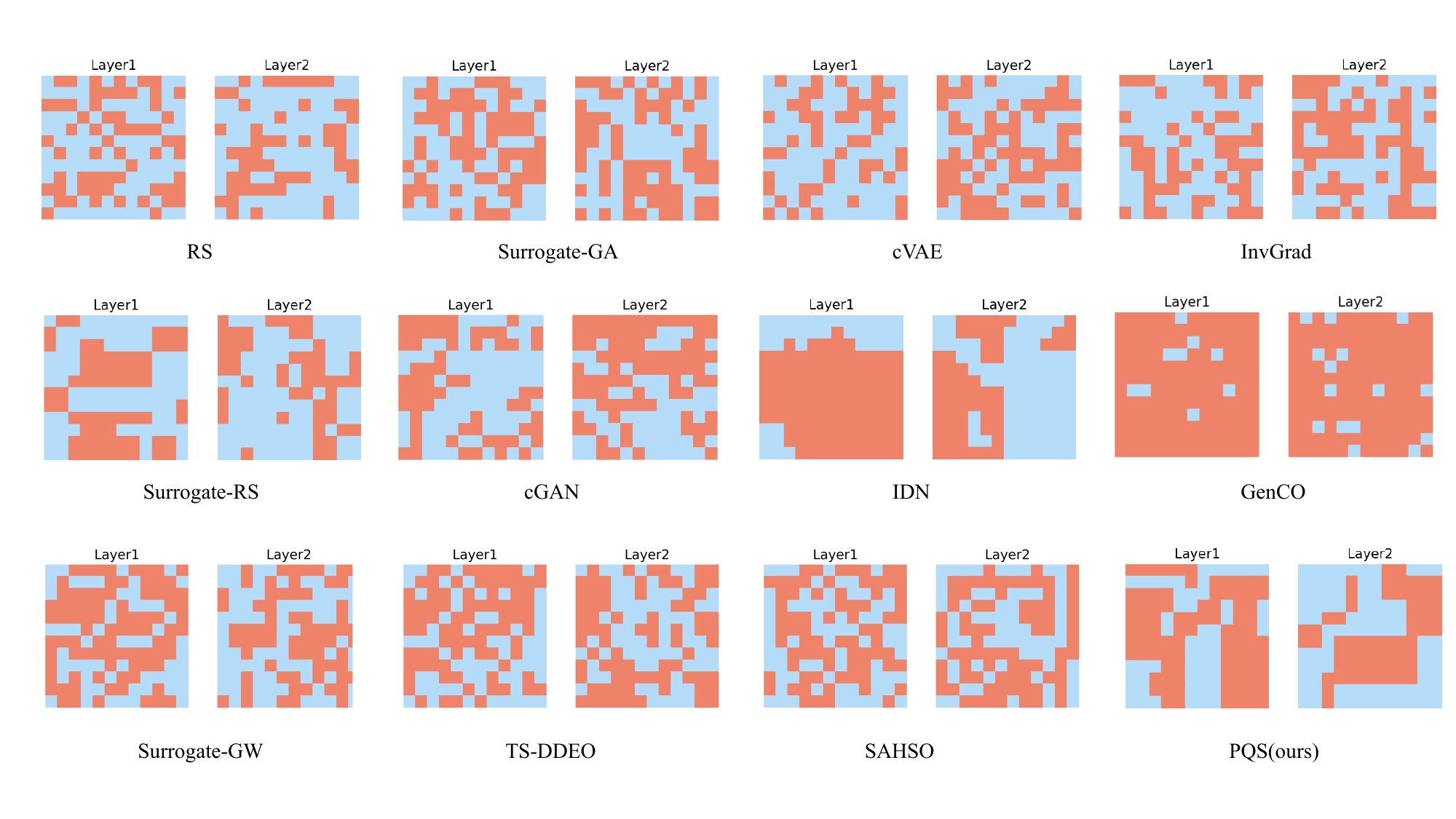}
  \caption{\refine{Optimized Electromagnetic Structures of Different Methods on Dual-layer Frequency Selective Surface.}}
  \label{fig:FSS_all_structures}
\end{figure}

\textbf{Comparisons of electromagnetic structures on Dual-layer Frequency Selective Surface}.
In Figure ~\ref{fig:FSS_all_structures}, we visualize the optimized electromagnetic structures searched by our proposed method and the baseline methods in the task of dual-layer frequency selective surface. It can be observed that, compared to the baseline methods, the structures designed by our approach exhibit a better adherence to physical priors, showcasing a more regular and manufacturable design.

\begin{figure}
  \centering
  \includegraphics[width=1\linewidth]
  {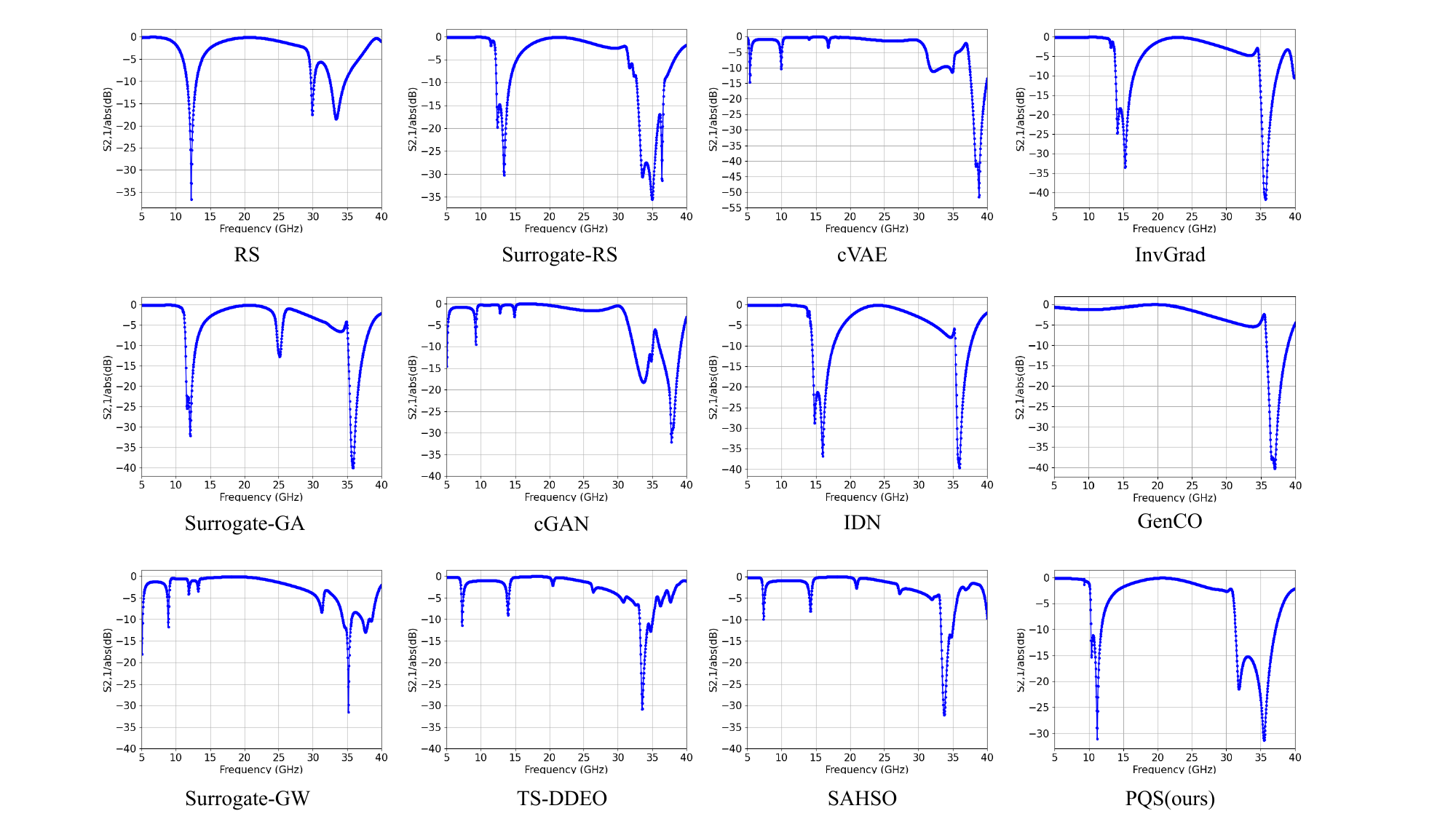}
  \caption{\refine{Simulated Results of Optimized Electromagnetic Structures on Dual-layer Frequency Selective Surface.}}
  \label{fig:FSS_all_simulations}
\end{figure}

\textbf{Comparisons of Simulated Results on Dual-layer Frequency Selective Surface}. In Figure ~\ref{fig:FSS_all_simulations}, We present the simulated results of the optimized electromagnetic structures for our methods and all baseline methods. From the results, it is evident that the optimized electromagnetic structures obtained through our proposed method exhibit lower values in the frequency ranges [31.5, 34.5] and [10.5, 15.5]. This indicates that our dual-layer frequency-selective surface performs better, providing empirical evidence for the effectiveness of our optimization algorithm.

\begin{figure}
  \centering
  \includegraphics[width=1\linewidth]
  {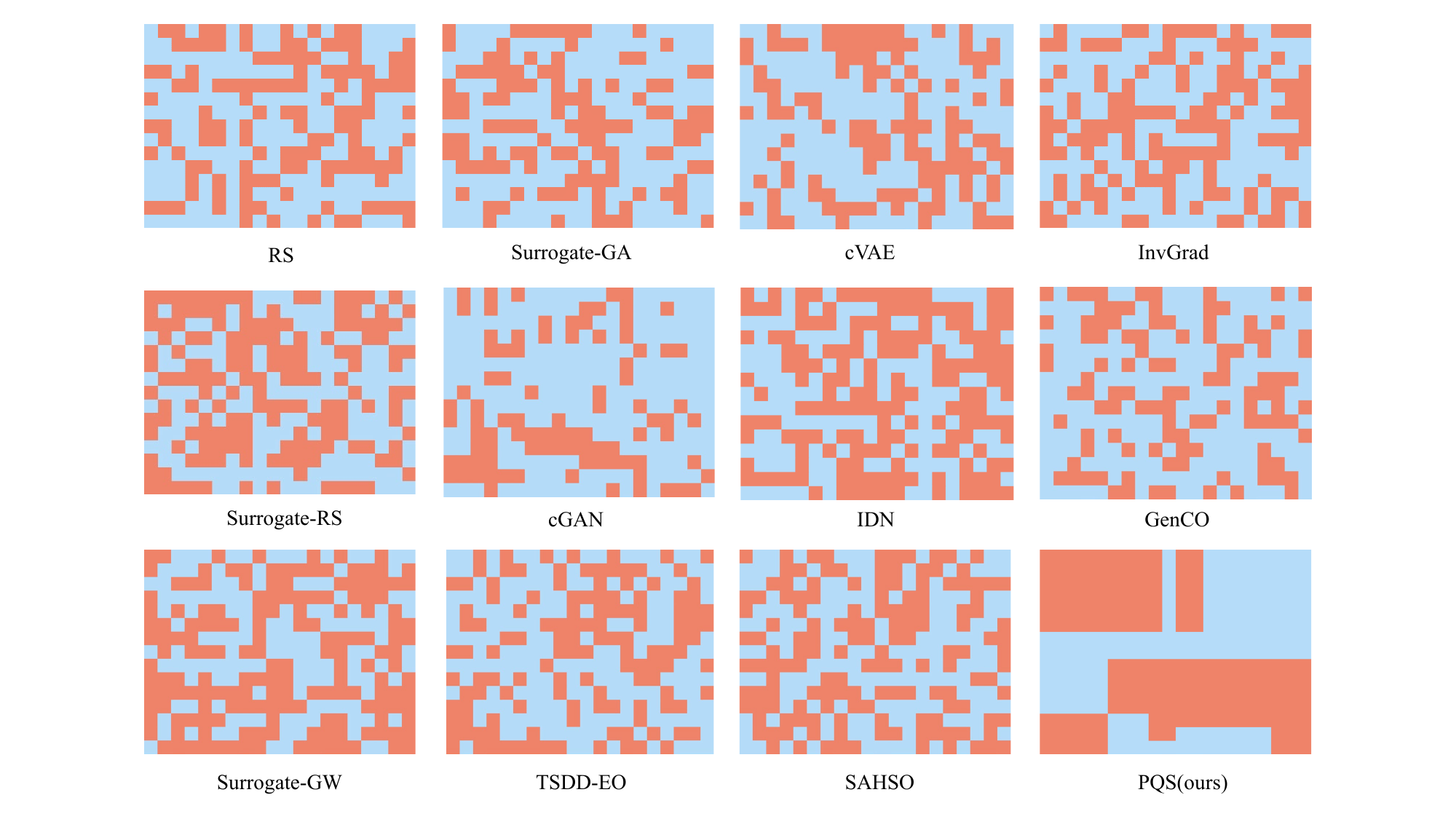}
  \caption{\refine{Optimized Electromagnetic Structures of Different Methods on High-gain Antenna.}}
  \label{fig:High-gain_all_mats}
\end{figure}

\textbf{Comparisons of electromagnetic structures on High-gain Antenna}. In Figure ~\ref{fig:High-gain_all_mats}, we visualize the optimized electromagnetic structures designed by our proposed method and the baseline methods in the task of high-gain antenna. It can be observed that, compared to the baseline methods, the structures designed by our approach also exhibit a better adherence to physical priors, showcasing a more regular and manufacturable design. This further demonstrates the strong generalization ability and robustness of our proposed method, proving its effectiveness across multiple real-world tasks.

\begin{figure}
  \centering
  \includegraphics[width=1\linewidth]
  {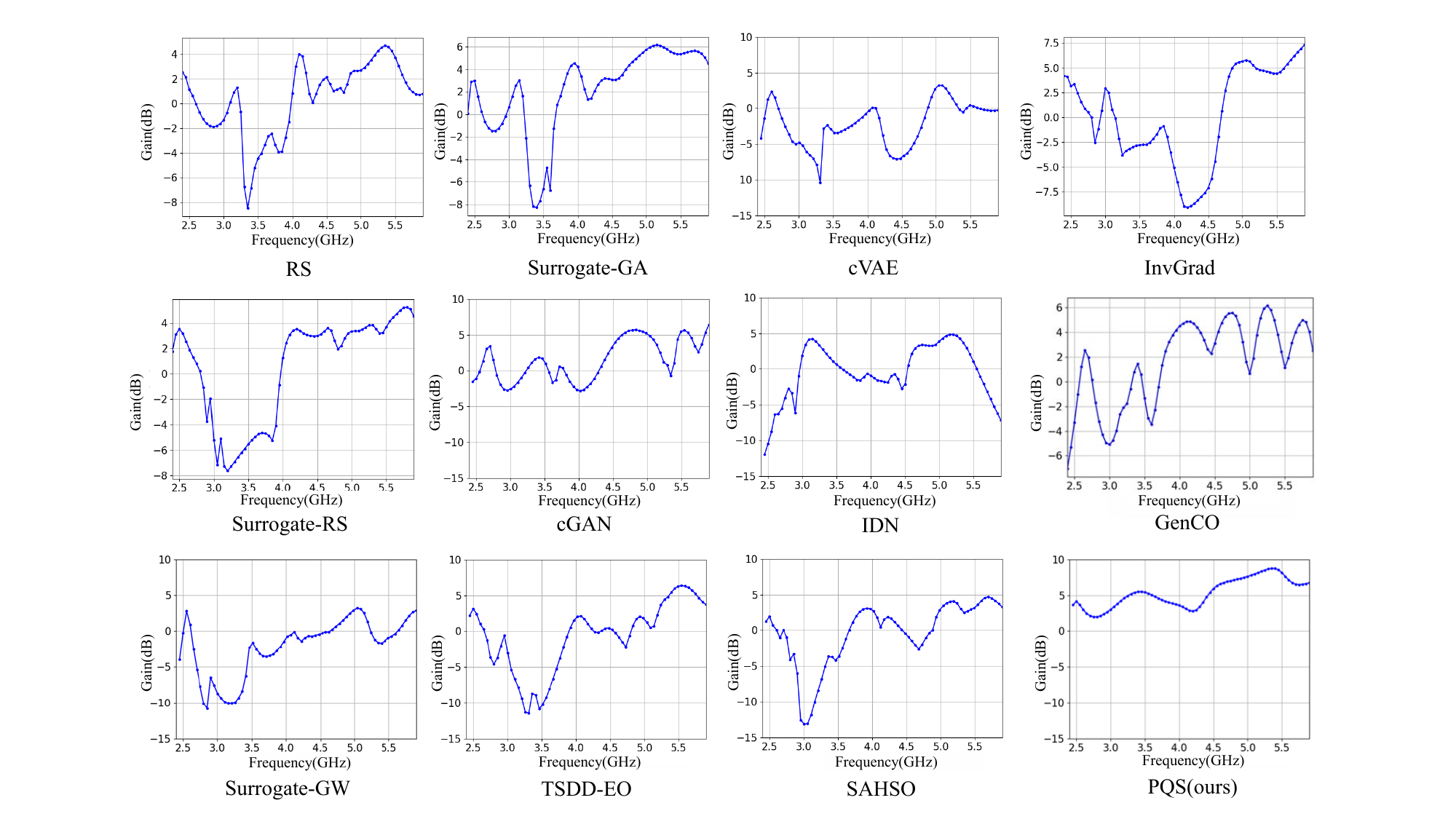}
  \caption{\refine{Simulated Results of Optimized Electromagnetic Structures on High-gain Antenna.}}
  \label{fig:High-gain_all_targets}
\end{figure}

\textbf{Comparisons of Simulated Results on High-gain Antenna}.  In Figure ~\ref{fig:High-gain_all_targets}, We further present the simulated results of the optimized electromagnetic structures for our methods and all baseline methods. The experiments further demonstrate that in the frequency ranges [2.45, 2.55] and [5.00, 6.00], the structures designed by our method significantly outperform all baseline methods, achieving substantial performance improvements in the target frequency bands.

\begin{figure}
  \centering
    \includegraphics[width=1\linewidth]{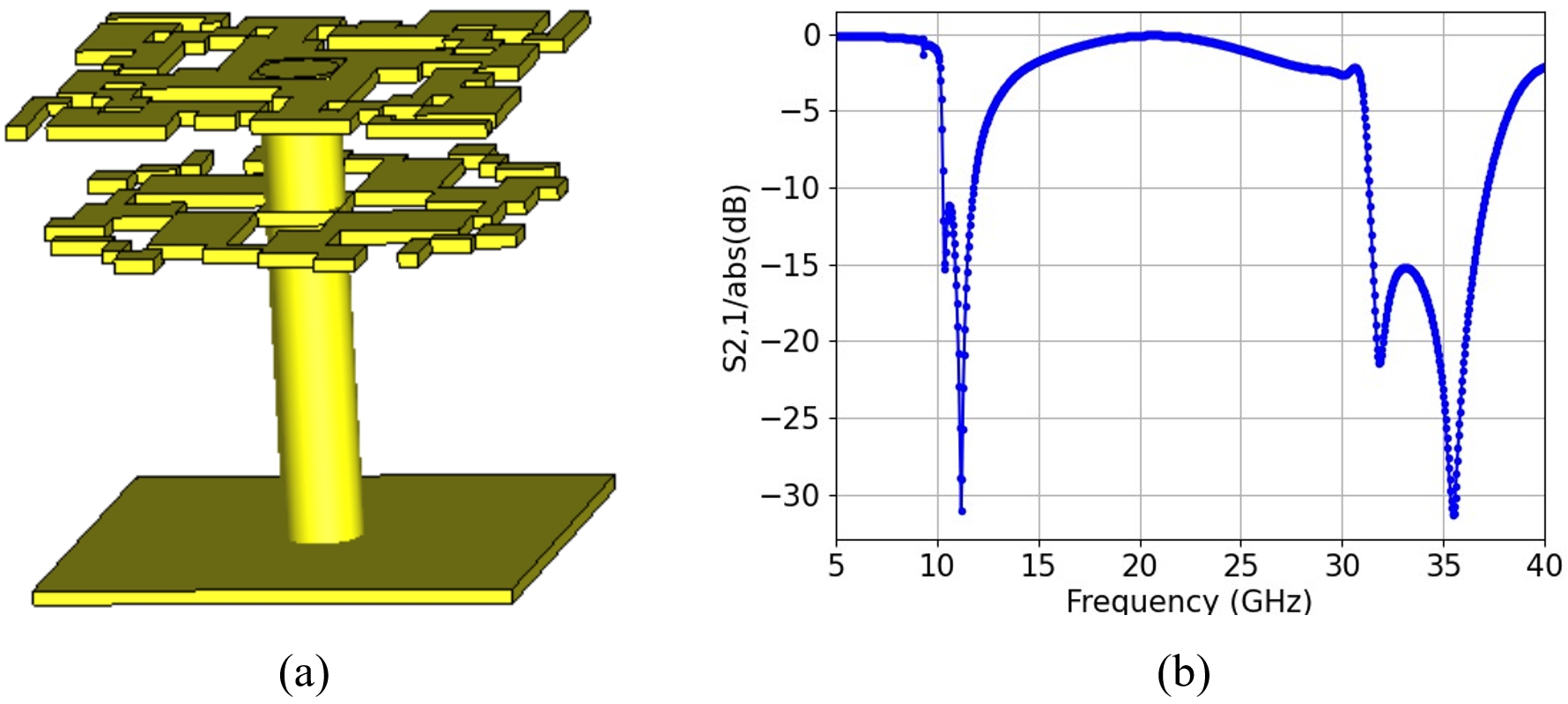}
    \vspace{-0.2in}
  \caption{(a) Optimized Electromagnetic Structures of Different Methods on Dual-layer Frequency Selective Surface. (b) Simulated Results of Optimized Electromagnetic Structures on Dual-layer Frequency Selective Surface.}
  \label{fig:ours_FSS}
  \vspace{-3mm}
\end{figure}

\textbf{Visualizations of Optimized Dual-layer Frequency Selective Surface}.We illustrate the optimized electromagnetic structures obtained through our proposed methodology and the reference methods in the High-gain Antenna task. In Figure ~\ref{fig:ours_FSS}, in contrast to the reference methods, the structures formulated by our approach demonstrate a heightened conformity to physical priors, presenting a more regular and manufacturable design.

\begin{figure}
  \centering
    \includegraphics[width=0.7\linewidth]{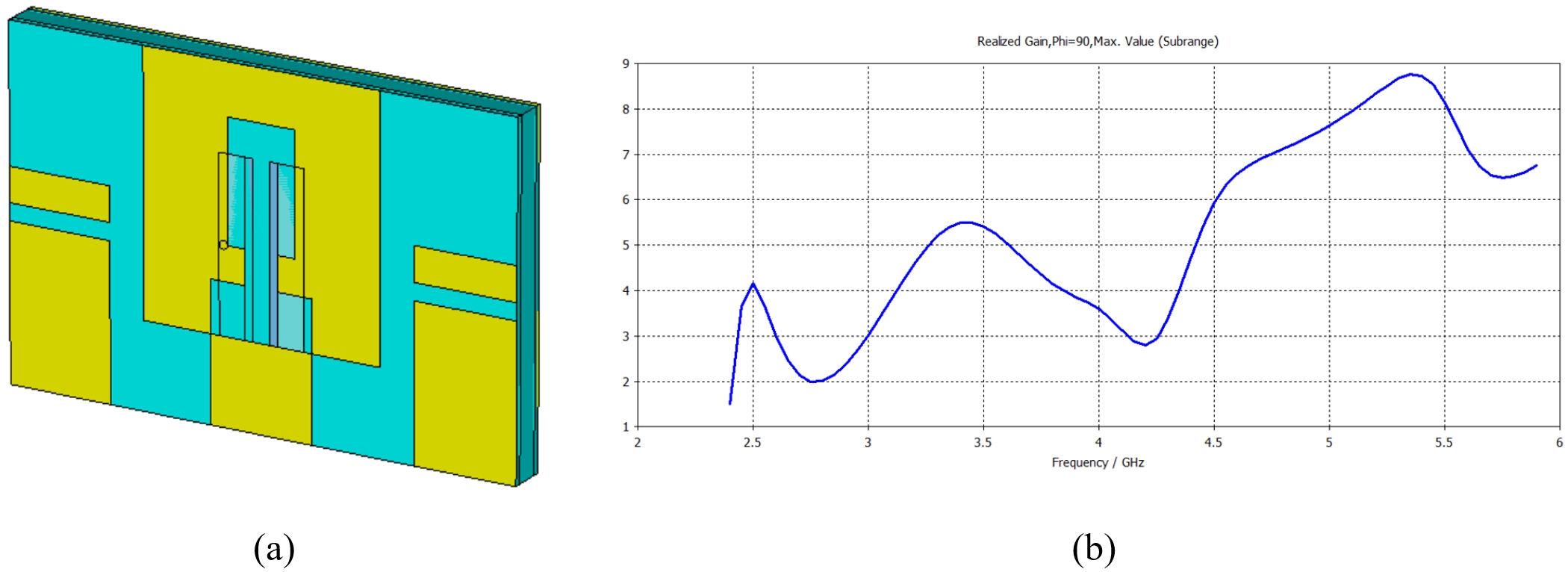}
    \vspace{-0.2in}
  \caption{(a) Optimized Electromagnetic Structures of Different Methods on the High-gain Antenna. (b) Simulated Results of optimized Electromagnetic Structures on High-gain Antenna.}
  \label{fig:ours_higngain}
  \vspace{-3mm}
\end{figure}

\textbf{Visualizations of Optimized High-gain Antenna}. In Figure ~\ref{fig:ours_higngain}, the results indicate that the optimized electromagnetic structures obtained through our proposed method exhibit higher values in the frequency ranges [2.45, 2.55] and [5.00, 6.00]. This implies that our high-gain antenna demonstrates superior performance, providing validation for the effectiveness of our algorithm.

\end{document}